\documentclass[11pt, a4paper, copyright, goog]{google}
\usepackage{etoolbox}       %
\usepackage{graphicx}  %
\usepackage[parfill]{parskip}
\usepackage{amssymb,amsmath,amsthm,bm}
\usepackage{mathtools}  %
\usepackage[dvipsnames]{xcolor}         %
\usepackage{thmtools} %
\usepackage{arydshln} %

\usepackage{colortbl}     %
\usepackage[table]{xcolor} %
\usepackage{booktabs}     %
\usepackage{multirow}     %
\usepackage{hyperref}
\usepackage{cleveref}
\usepackage{pifont}%
\newcommand{\cmark}{\ding{51}}%
\newcommand{\xmark}{\ding{55}}%

\usepackage{bm}
\usepackage[dvipsnames]{xcolor}  %
\usepackage{subcaption}
\usepackage{wrapfig,lipsum}
\usepackage{algorithm,algorithmicx,algpseudocode}
\usepackage{nicematrix}

\usepackage{siunitx}
\usepackage[normalem]{ulem}
\robustify\bfseries
\robustify\uline
\def\Uline#1{#1\llap{\uline{\phantom{#1}}}}
\newrobustcmd\Bf{\DeclareFontSeriesDefault[rm]{bf}{b}\bfseries}

\usepackage{listings}
\lstdefinestyle{pseudocode}{
    basicstyle=\ttfamily\small,
    keywordstyle=\bfseries,
    commentstyle=\itshape,
    showstringspaces=false,
    columns=fullflexible,
    keepspaces=true,
    frame=single,
    mathescape=true,
    escapeinside={(*@}{@*)}
}

\usepackage[authoryear, sort&compress, round]{natbib}
\DeclareMathOperator{\softmax}{softmax}
\DeclareMathOperator{\attention}{attention}
\DeclareMathOperator{\mamba}{Mamba-2}
\DeclareMathOperator{\gdn}{GDN}

\newcommand{\bmk}{\bm{k}}
\newcommand{\bmo}{\bm{o}}
\newcommand{\bmq}{\bm{q}}
\newcommand{\bmv}{\bm{v}}
\newcommand{\bmx}{\bm{x}}

\newcommand{\bmA}{\bm{A}}

\newcommand{\bmI}{\bm{I}}
\newcommand{\bmK}{\bm{K}}
\newcommand{\bmL}{\bm{L}}
\newcommand{\bmM}{\bm{M}}
\newcommand{\bmO}{\bm{O}}
\newcommand{\bmQ}{\bm{Q}}
\newcommand{\bmS}{\bm{S}}

\newcommand{\bmV}{\bm{V}}
\newcommand{\bmW}{\bm{W}}
\newcommand{\bmX}{\bm{X}}
\newcommand{\bmY}{\bm{Y}}

\newcommand{\bmbeta}{\bm{\beta}}
\newcommand{\bmGamma}{\bm{\Gamma}}

\newcommand{\mixer}{{\bmM}}

\uselogo{} 

\title{Multi-Mixer Models: Flexible Sequence Modeling with Shared Representations}

\correspondingauthor{kyl2@cs.cmu.edu}

\reportnumber{n/a} %

\newcommand{\antelope}{\textsc{Oryx}}

\author[1,*]{Kevin Y. Li}
\author[2]{Asher Trockman}
\author[2]{Ananda Theertha Suresh}
\author[2]{Ziteng Sun}

\affil[1]{Carnegie Mellon University}
\affil[2]{Google Research}
\affil[*]{Work done while the author was at Google Research.}

\begin{abstract}
Softmax attention is the cornerstone of modern large language models.
However, its memory requirements scale linearly with sequence length, and its compute requirements scale quadratically.
Linear recurrent models, such as linear attention and state space models, have become widely studied as alternatives to softmax attention due to their linear compute and constant memory requirements.
While these sub-quadratic token mixing methods (mixers) achieve promising efficiency gains and competitive results on a wide range of benchmarks, current linear recurrent models still lag behind on tasks that require long-context retrieval or in-context learning.
A growing body of work studies hybrid architectures that attempt to mitigate these trade-offs by statically interleaving or merging attention and recurrent blocks. 
In this work, we explore a new axis of developing hybrid models: across the token sequence. We propose \antelope, a hybrid model that can, throughout a sequence, flexibly switch between different mixers, e.g., quadratic attention, for rich context utilization, and linear recurrences, for efficient generation.
\antelope{} ties at least 90\% of its parameters across mixers, enabling attention and recurrent modes to operate over shared internal representations.
We validate our design with Mamba-2 and Gated DeltaNet variants, up to 1.4B models.
Under fixed token budgets and a mixed-training strategy, \antelope{} achieves comparable or better performance than its single-mixer baselines.
At the 1.4B scale, all instances of \antelope{} outperform their respective baselines by at least \textbf{0.7} percentage points on averaged language modeling tasks. On retrieval tasks, \antelope{} achieves performance comparable to the Transformer baseline even when processing only a tiny fraction ($<$10\%) of the tokens in the attention mode.
These results suggest that attention and linear recurrent models can share internal representations, and motivate sequence-axis hybridization as a promising direction.
\end{abstract}

\begin{document}
\maketitle

\section{Introduction}

The Transformer architecture and its core softmax attention mechanism remain the dominant foundation for modern large language models (LLMs).
However, softmax attention maintains a key-value (KV) cache of all previous tokens and queries the full cache at each step, incurring quadratic compute and linear memory requirements in sequence length.
These costs have motivated
the recent proliferation of sub-quadratic alternatives, in particular, linear recurrent models, such as linear attention~\citep{katharopoulos2020transformersrnnsfastautoregressive,yang2025gateddeltanetworksimproving} and state-space models (SSMs)~\citep{gu2024mambalineartimesequencemodeling,dao2024transformersssmsgeneralizedmodels}.\footnote{\,In this work, we use attention to refer to the canonical quadratic softmax attention mechanism, and use linear to refer to the general class of linear recurrent mechanisms. 
}
These linear models are characterized by their constant-size recurrent state, which is updated after each token, and linearly-scaling compute. While they have demonstrated promising results in many settings, pure recurrent approaches still lag behind on tasks that require strong retrieval or in-context learning abilities~\citep{waleffe2024empiricalstudymambabasedlanguage,arora2025simplelinearattentionlanguage}.
Such trade-offs have motivated the development and deployment of hybrid architectures, which combine these linear layers with softmax attention to balance performance and efficiency~\citep{waleffe2024empiricalstudymambabasedlanguage,kimiteam2025kimilinearexpressiveefficient,nvidia2025nvidianemotron3efficient,qwen3technicalreport}.

Existing hybrid models fall under two main paradigms: inter-layer designs, which interleave softmax attention and linear layers, and intra-layer designs, which fuse the output of attention and linear mechanisms within a single layer or block.
For both, the computational cost per token and capabilities are largely defined by the predetermined architecture.
In practice, however, the required modeling capabilities and desired trade-off vary by task: retrieval tasks may benefit from richer attention-based context utilization, whereas standard language generation tasks may be adequately served by linear computation.
This suggests a complementary form of hybridization: instead of choosing a fixed mixture of mechanisms, \emph{can a model operate with different mixers throughout the sequence?}

\begin{figure*}[t]
\centering
\includegraphics[width=\linewidth]{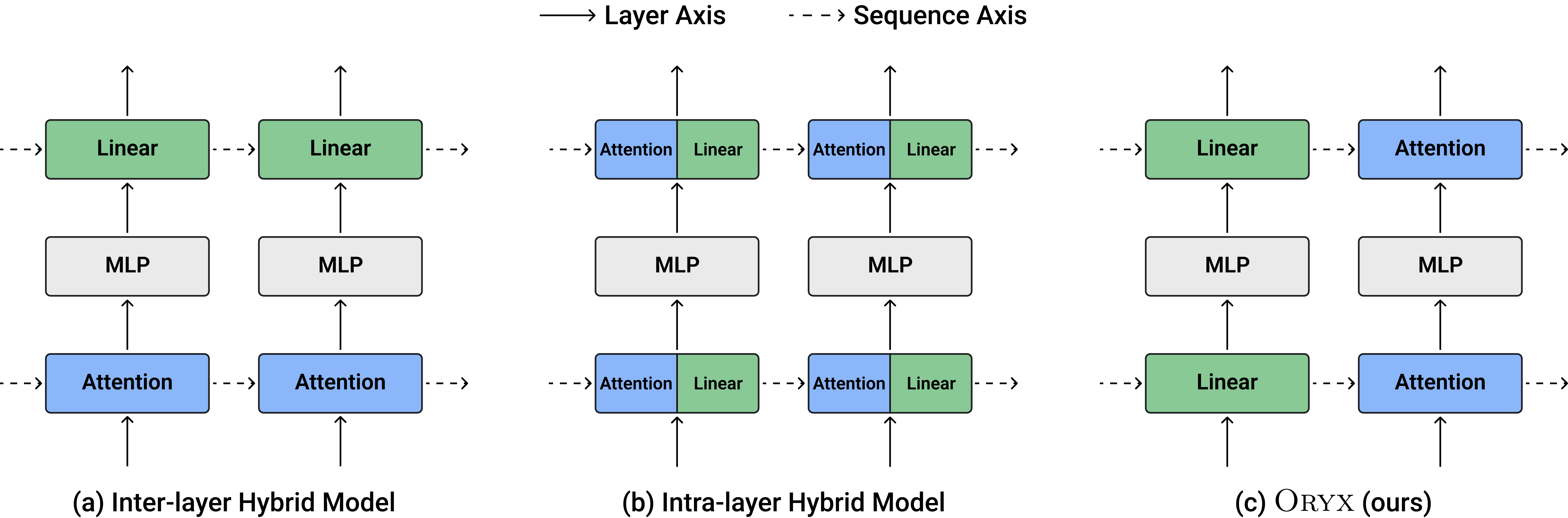}
\caption{
Comparison of different hybrid architectures.
(a) Inter-layer hybrid models interleave different mixers along the layer axis;
(b) Intra-layer hybrid models fuse different mixers within a single layer;
(c) \antelope~is a sequence-axis hybrid model that can switch between different mixers across the sequence, allowing different segments of the input to be processed by varying mechanisms.
}
\label{fig:hybrid_comparison}
\end{figure*}

In this work, we propose \antelope, a sequence-axis hybrid architecture that supports operating in \emph{both} quadratic-attention and linear-recurrent regimes throughout a sequence (\Cref{fig:hybrid_comparison}), trained by a chunked mixed-mode strategy that enables flexible switching during deployment (\Cref{sec:method}).
A key challenge in supporting mode switching is that attention and recurrent mixers parameterize and update their state differently.
To bridge these mechanisms, \antelope{} layers maintain a KV cache and a linear recurrent state, updating both jointly at each timestep with key-value pairs obtained from shared weights across mixer types.
Thus, both mechanisms update states from a shared representation space, avoiding separate state-update features for each mixer.
Consequently, when the mixer mechanism switches, the new selected mixer can continue from a compatible state accumulated over all preceding tokens.
This selection flexibility during inference could enable compute allocation at the prompt or token level.
For example, reasoning traces could be generated with the linear mixer for lower-latency, and the answer could use the attention mode for better retrieval and summarization.

We instantiate this design with  Mamba-2~\citep{dao2024transformersssmsgeneralizedmodels} and Gated DeltaNet (GDN)~\citep{yang2025gateddeltanetworksimproving} as the linear mixer, with both variants \emph{sharing more than $90\%$ of parameters across mixer modes}.
Although our experimental validation focuses on Mamba-2 and GDN, we underscore that our overall design is not specific to these choices and can extend to other mixers that admit comparable key-value associations.
Our multi-mixer models can switch between attention and linear sequence processing while incurring little to no degradation in output quality (\Cref{fig:mode-switch-1B}), and
\emph{under matched parameters and training token budgets}, \antelope{} remains competitive with pure softmax attention, Mamba-2, and GDN baselines on language modeling performance across scale.
At the 1.4B scale, both attention and linear modes of Mamba-2 and GDN variants of \antelope{} outperform their respective baselines by \emph{more than $0.7$} percentage points on average across downstream language modeling evaluations (\Cref{tab:downstream_evaluation}).
On a variety of real-world retrieval tasks and synthetic retrieval tasks,
\antelope{} is able to achieve comparable performance compared to the Transformer baseline with $<$10\% of the tokens processed in the attention mode.
Using this mixed-inference mode, \antelope{} \emph{significantly surpasses the linear baselines} by a margin of at least 8.6 percentage points on real-world retrieval tasks and at least 38.6 percentage points on needle-in-a-haystack (NIAH) tests (\Cref{tab:retrieval_cross_mode}), demonstrating that \antelope{} is a promising method for improving the retrieval capabilities of linear models.
Our findings suggest that \emph{attention and linear recurrent mechanisms can share similar representations} despite differing largely in methodology, opening new opportunities to study their interplay. 

We first introduce preliminary background on sequence mixers and their connections in \Cref{sec:background}.
We use their commonality to motivate the design of the multi-mixer \antelope{} block in \Cref{sec:method} and highlight the main architectural components.
\Cref{sec:results} evaluates the language modeling and mode-switching ability of the \antelope{} model and ablates core design choices.

\section{Preliminaries}
\label{sec:background}
\paragraph{Notation.}
We use the term \textbf{mixer} to denote a transformation that mixes information along a particular axis of the input.
A \textbf{sequence mixer} (e.g., softmax attention) mixes information across the sequence dimension, whereas a \textbf{channel mixer} (e.g., an MLP) mixes across the feature or channel dimension.
We focus on sequence mixer designs in this paper and use the term mixer to refer to sequence mixer unless otherwise specified.
We denote scalars by non-bold symbols (e.g., $a $), vectors by bold lowercase symbols (e.g., $\bmq $), and matrices by bold uppercase symbols (e.g., $\bmX, \bmW$).
Subscripts index the first axis by default, while superscripts are reserved for identifiers.

\paragraph{Shared Key-Value Association View.}
While softmax attention~\citep{vaswani2023attentionneed} and modern linear models, e.g., state-space models~\citep{dao2024transformersssmsgeneralizedmodels}, linear attention~\citep{katharopoulos2020transformersrnnsfastautoregressive}, and fast-weight programmers~\citep{schlag2021lineartransformerssecretlyfast,yang2025gateddeltanetworksimproving}, differ in how they store and update their state, they can all be unified under the associative memory view~\citep{wang2025testtimeregressionunifyingframework, liu2024longhornstatespacemodels}.
Under this view, each mechanism maintains a memory of key-value associations and uses queries to retrieve relevant values from that memory.
Their core representations, namely the queries, keys, and values, are obtained through linear projections of the input, parameterized by weight matrices $\bmW^Q, \bmW^K$, and $\bmW^V$. For input $\bmx \in \mathbb{R}^{1 \times D}$, we have 
\begin{equation*}
    \bmq  = \bmx  \bmW^Q,\quad
    \bmk  = \bmx  \bmW^K,\quad
    \bmv  = \bmx  \bmW^V,
\end{equation*}
where $\{\bmW^Q, \bmW^K\} \in\mathbb{R}^{D\times D_k}, \bmW^V\in\mathbb{R}^{D\times D_v}$. For input sequence $\bmX  := [\bmx_1; \bmx_2; \ldots; \bmx_T ] \in \mathbb{R}^{T\times D}$, we use $\bmQ  = [\bmq_1; \bmq_2; \ldots; \bmq_T ]$ to denote all queries, and define $\bmK $ and $\bmV $ similarly.
This viewpoint provides the basis for the tied-projection design in \Cref{sec:method}.

\paragraph{Softmax Attention.}
For causal softmax attention, the output $\bmo_t \in \mathbb{R}^{1 \times D_v}$ at timestep $t$ is a weighted aggregation of past values $\bmV_{\le t}$ according to the softmax-normalized similarity between the current query $\bmq_t$ and past keys $\bmK_{\le t}$: $\softmax\!\left(\bmq_t \bmK_{\le t} ^\top\right)\bmV_{\le t}$.
\footnote{\,We ignore positional encodings and other complementary components, e.g., head structure, head dimension normalization, for clarity.}
In parallel form, it is expressed as
\begin{equation*}
    \bmO  = \text{Mixer}_{\attention}(\bmQ , \bmK , \bmV ) :=  \softmax\!\left(\bmL^{U} + (\bmQ \bmK ^\top) \right)\bmV ,
\end{equation*}
where 
$\bmL^{U}\in\mathbb{R}^{T\times T}, \bmL^U_{ij} = -\infty \cdot \mathbb{I}[i < j]$ is the causal mask.
Its state, the KV cache, stores the key and value vectors $\bmk_t, \bmv_t$ from all previous time steps and therefore grows linearly with sequence length.
The mixer output $\bmO$ is passed through the output projection $\bmW^O \in \mathbb{R}^{D_{v}\times D} $ to get the final block output in the original input dimension,
$\bmY  = \bmO  \bmW^O$.
As this output projection step is present in other models, we will ignore it for clarity from this point onward.

\paragraph{Linear Recurrent Neural Networks (RNNs).} In contrast to softmax attention, linear RNNs, which encompass linear attention variants and modern state space models (SSMs), are grounded in their recurrent structure.
A key characteristic of these models is their fixed-size states, which enable runtime and memory efficiency.
Despite not having an explicit cache of key-value pairs, the states \emph{are} updated with key-value associations at each timestep.
In general, a structured transition matrix $\bmA_t \in \mathbb{R}^{D_k \times D_k}$ adjusts the prior state $\bmS_{t-1} \in \mathbb{R}^{D_k \times D_v}$ while the current key-value interaction is incorporated via an outer-product.
The output is determined using a simple readout with the current query.
\[
    \bmS_t = \bmA_t \bmS_{t-1} + \bmk_t^\top \bmv_t,
    \qquad
    \bmo_t = \bmq_t \bmS_t,
\]
Thus, its state and output can still be expressed through query, key, and value representations.

\paragraph{Mamba-2.} The discretized Mamba-2 SSM~\citep{dao2024transformersssmsgeneralizedmodels} is one instantiation of a linear RNN that uses a scalar times identity transition, $\bmA_t = \alpha_t \bmI$, where $\alpha_t$ is an input-dependent decay factor.
\footnote{\,In Mamba-2, queries, keys, and values are referred to as $C, B, x$ respectively, but we utilize attention terminology to draw similarities.
We also ignore the discretization parameters and the tied nature of $\alpha_t$ and $\bmv_t$ in this section for clarity.}
Its parallel form relies on decay-based lower-triangular mask $\bmGamma$ applied with a Hadamard product ($\circ$), which draws connections to attention variants and highlights the value retrieval mechanism. %
\begin{align*}
  \bmO  = \text{Mixer}_{\mamba}(\bmQ , \bmK , \bmV, \bmX) := \left(
\bmGamma 
\circ \left(\bmQ  \bmK ^\top\right)
\right)\bmV ,\quad
\bmGamma = {\scriptstyle
\begin{bmatrix}
1 \\
\alpha_2 & 1 \\
\alpha_3 \alpha_2 & \alpha_3 & 1 \\
\vdots & & & \ddots
\end{bmatrix}}
\end{align*}

\paragraph{Gated DeltaNet.}
Fast-weight programmers~\citep{schlag2021lineartransformerssecretlyfast, yang2025parallelizinglineartransformersdelta}, such as Gated DeltaNet~\citep{yang2025gateddeltanetworksimproving}, can also be viewed under this lens.
While the memory is queried the same way, the gated delta update rule enables a more expressive state update mechanism
\begin{align*}
    \bmS_t  = \left(\alpha_t\left(\bmI - \beta_t \bmk_t^\top \bmk_t \right)  \right)\bmS_{t-1} + \beta_t \bmk_t^\top \bmv_t, \quad
    \bmo_t  = \bmq_t \bmS_t,
\end{align*}
where $\alpha_t, \beta_t$ are both data-dependent scalars.
Like other sub-quadratic alternatives, it also retains a parallel representation.
Reusing the decay mask $\bmGamma$ from the Mamba-2 formulation, we have
\begin{align*}
\bmO
&=
\operatorname{Mixer}_{\gdn}
(\bmQ,\bmK,\bmV;\bmGamma,\bmbeta)
:=
\left(
\bmGamma \circ \left(\bmQ\bmK^\top\right)
\right)
\left[
\bmI
+
\operatorname{strictLower}\!\left(
\operatorname{diag}(\bmbeta)
\left(
\bmGamma \circ \left(\bmK\bmK^\top\right)
\right)
\right)
\right]^{-1}
\operatorname{diag}(\bmbeta)
\bmV .
\end{align*}

While these sequence mixers differ in how they store (e.g., KV cache or fixed state), normalize (e.g., softmax, decay mask), and update (e.g., delta rule) key-value associations, their common query-key-value interaction structure provides a useful lens for connecting softmax attention and linear models.
This connection motivates the design of our \antelope{} block in \Cref{sec:method}, where tied projections are used to update both attention and recurrent states from shared representations. %

\section{Designing the Multi-Mixer \antelope{} Shared Block}
\label{sec:method}
\begin{figure}[!t]
    \centering
    \includegraphics[width=0.9\linewidth]{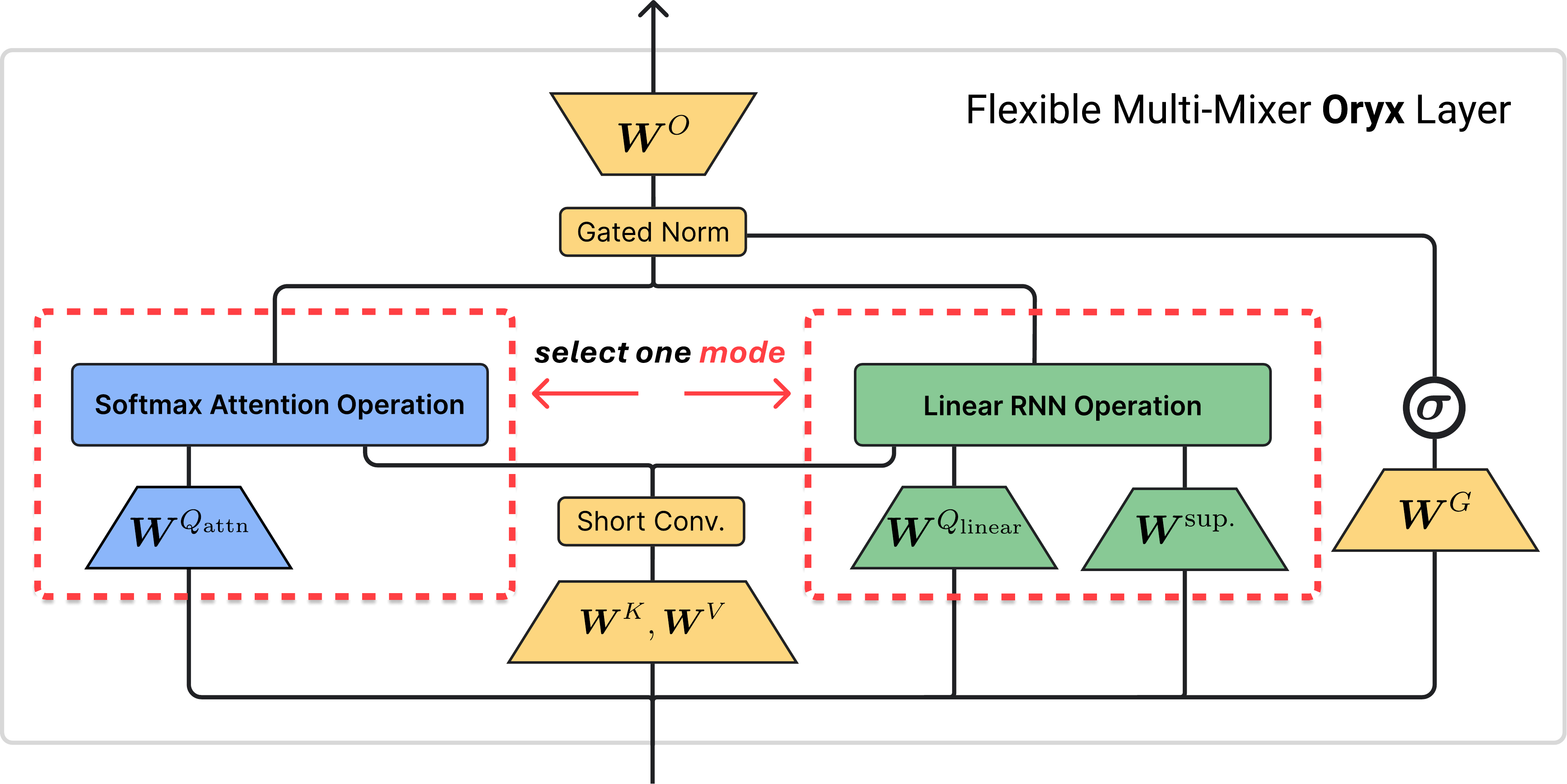}
    \caption{
    General \antelope{} block.
    The block shares the core representations between softmax attention and a linear recurrent mechanism through tied key-value projections and uses additional components critical to the linear mechanism's performance, e.g., the short convolution and gate.
    During a forward pass, the shared key-value representation updates both the KV cache and linear state, allowing either the softmax attention mixer or linear mixer to be chosen as the mode of operation at each timestep.
    }
    \label{fig:arch}
\end{figure}

In this section, we describe the sequence-mixing \antelope{} shared block that supports operating in both softmax attention and linear recurrent mechanisms.
The resulting block maintains compatible attention and recurrent states and includes architectural components core to the modeling abilities of each mechanism while sharing most of its parameters across mixers (\Cref{fig:arch}).
We instantiate the linear mixer as Mamba-2 and Gated DeltaNet (GDN) for concreteness, but our design choices \emph{can} be applied with other linear models with similar query-key-value interactions.%

\paragraph{Shared Key-Values and Mixer-Specific Queries.}

Motivated by the key-value associative memory view (\Cref{sec:background}), \antelope{} ties the key and value projections across mixer modes, enabling one set of representations, calculated from one forward pass, to update both the attention KV cache and linear recurrent state.
While the query projections can also be shared, our empirical results suggest that tying all three core representations across mixers hinders model performance (see details in \Cref{sec:results:subsec:ablations}), and thus, we use unique weights for each mixer's query projection for stronger performance.
We conjecture that the differences in their state update and output rules may require different query vectors to extract mode-specific crucial information, a hypothesis supported by empirical results.

\paragraph{Incorporating Additional Linear Model Components.}
\antelope{} incorporates the short convolution, multiplicative gate, and pre-output projection normalization --- common components in current linear models --- to retain these important inductive biases.
In our implementation, the short convolution is applied to only the shared keys and values before they are passed into the selected $\text{Mixer}_{\mixer}$; the queries are not convolved due to being mixer-specific.

For a selected mixer mode $\mixer \in \{\attention, \mamba, \gdn\}$, the block computes\footnote{\,Rotary embeddings, the short convolution, etc., are abstracted away in the $\text{Mixer}$ class. \Cref{sec:additional-ablations} details the exact architecture for each \antelope{} variant.}
\begin{align*}
    \bmO & = \text{Mixer}_{\mixer}(\bmX\bmW^{Q_{\mixer}}, \bmX\bmW^K, \bmX\bmW^V, \bmX; \bmW^\text{sup.}),
\end{align*}
where $\bmW^\text{sup.}$ denotes the additional mixer-specific supporting parameters, either data-dependent or independent, such as discretization or delta-rule parameters.
The mixer output is gated element-wise and normalized before the final shared output projection
\begin{align*}
    \bmY & = \text{GatedRMSNorm}\left(\bmO, \sigma(\bmX\bmW^G)\right) \bmW^O,
\end{align*}
where $\sigma$ is an activation function, usually SiLU~\citep{hendrycks2023gaussianerrorlinearunits}, and $\bmW^G \in \mathbb{R}^{D\times D_v}$ is the shared gate projection.
The exact normalization and supporting parameters depend on the linear mechanism used;
we detail the specifics of our variants in \Cref{sec:additional-ablations}.
Most parameters, including $\bmW^K$, $\bmW^V$, $\bmW^G$, and $\bmW^O$, are shared across mixers.
Thus, these shared representations can be calculated with one shared forward pass.
The general block structure is visualized in \Cref{fig:arch}, and an abstracted version of the pseudocode can be found at Listing~\ref{lst:oryx_clean_pseudocode}.
\vspace{-0.5em}
\paragraph{Maintaining Compatible States and Head Structures.}
At each timestep, \antelope{} computes shared key and value representations and uses them to update both the attention KV cache and linear recurrent state.
While the selected mixer determines the block output, the joint update allows both states to retain the same token history.
We note that while both mixer states are maintained, only the updated state of the selected mixer is needed for output computation.
However, one issue is that different mixers often use differing head structures.
For instance, modern Transformers use attention in a multi-head (MHA) or grouped-query head (GQA) structure, while Mamba-2 adopts a multi-value (MVA) structure.
To enable the effective sharing of weight projections, we use the same head structure across mixers, matching that of attention, i.e., MHA in our experiments.
Despite this constraint applied on the linear mixers, we find that they remain empirically effective for modeling.
\vspace{-2em}
\paragraph{Chunked Mixed-Mode Training.}
To enable robust mode switching capabilities at inference time, we train \antelope{} with chunked mixed-mode training (see ablation in \Cref{sec:results:subsec:ablations}).
Each training sequence is partitioned into fixed-length chunks, e.g., of 128 tokens, and each chunk is randomly assigned a mixer mode, e.g., attention or linear mode.
All \antelope{} blocks use the same chunk assignment, and our experiments find that a 1:3 attention-to-linear chunk training ratio balances the performance.

\begin{table}[!t]
    \small
    \centering
    \caption{
    Downstream language modeling evaluations on parameter-matched models trained with 100B FineWeb-Edu tokens.
    We compare baseline models, \antelope{}-TM (Transformer/Mamba-2), and \antelope{}-TG (Transformer/Gated DeltaNet) at each parameter scale.
    Best results for each size are \textbf{bolded}, and second best are \underline{underlined}.
    The \antelope{} results reported below use the same mixer for the entire model and sequence.
    Under a fixed training token budget, both modes of our dual mixer \antelope{} model achieve comparable or better performances than that of single mixer baselines.
    }

    \resizebox{\textwidth}{!}{
    \sisetup{
        detect-weight,
        table-align-text-post=false,
        table-space-text-post={*}
    }
    \begin{tabular}{l l l *{19}{S[table-format=2.1]}}
        \toprule
        & \textbf{Family} & \textbf{Mixer} & {LAMB.} & {LAMB.} & {HellaS.} & {PIQA} & {Arc-E} & {Arc-C} & {WinoGr.} & {OBQA} & {\textbf{Avg.}} \\
        & & & {ppl $\downarrow$} & {acc $\uparrow$} & {acc\_n $\uparrow$} & {acc $\uparrow$} & {acc $\uparrow$} & {acc\_n $\uparrow$} & {acc $\uparrow$} & {acc $\uparrow$} & {acc $\uparrow$} \\
        \midrule
        \multirow{7}{*}{\rotatebox[origin=c]{90}{\textbf{130M}}}
        & \multirow{3}{*}{Baseline} & Transformer & 42.6 & 32.3 & 39.2 & 66.8 & 58.4 & \Uline{28.9} & 51.1 & 19.4 & 42.3 \\
        & & Mamba-2 & 41.5 & 29.9 & 40.0 & 67.1 & \Uline{60.0} & 27.9 & 52.6 & \Uline{22.8} & 42.9 \\
        & & Gated DeltaNet & \Bf 36.5 & 32.2 & \Bf 40.5 & \Bf 68.4 & \Bf 62.7 & 28.7 & 51.6 & 22.0 & \Bf 43.7 \\
        \addlinespace[0.6ex]
        \cdashline{2-12}
        \addlinespace[0.6ex]
        & \multirow{2}{*}{\antelope{}-TM} & Transformer & \Uline{38.2} & \Bf 34.3 & 39.3 & \Uline{67.7} & 58.7 & 28.1 & \Bf 54.0 & 22.4 & \Uline{43.5} \\
        & & Mamba-2 & 40.3 & 31.1 & 39.8 & 67.4 & 59.0 & \Bf 29.0 & \Uline{53.7} & \Bf 23.2 & 43.3 \\
        \addlinespace[0.6ex]
        \cdashline{2-12}
        \addlinespace[0.6ex]
        & \multirow{2}{*}{\antelope{}-TG} & Transformer & 39.3 & \Uline{32.7} & 39.9 & 67.3 & 59.7 & 28.6 & 50.9 & 21.4 & 42.9 \\
        & & Gated DeltaNet & 39.0 & 31.5 & \Uline{40.4} & 67.1 & 59.3 & 27.9 & 48.6 & 20.8 & 42.2 \\
        \midrule

        \multirow{7}{*}{\rotatebox[origin=c]{90}{\textbf{380M}}}
        & \multirow{3}{*}{Baseline} & Transformer & 19.5 & 41.1 & 51.0 & 70.8 & 68.2 & 33.6 & 55.7 & 23.8 & 49.2 \\
        & & Mamba-2 & 18.3 & 41.3 & \Bf 51.8 & \Bf 72.1 & 68.5 & \Uline{35.2} & 56.6 & 27.0 & 50.4 \\
        & & Gated DeltaNet & \Bf 16.5 & \Bf 42.4 & 51.4 & 71.2 & 68.6 & 34.6 & 55.3 & \Uline{27.2} & 50.1 \\
        \addlinespace[0.6ex]
        \cdashline{2-12}
        \addlinespace[0.6ex]
        & \multirow{2}{*}{\antelope{}-TM} & Transformer & \Uline{17.8} & 41.9 & 51.2 & 71.2 & \Bf 71.0 & 35.1 & 57.0 & 26.8 & \Uline{50.6} \\
        & & Mamba-2 & 18.5 & \Uline{42.3} & 51.0 & 71.3 & \Uline{70.5} & \Bf 36.3 & \Uline{57.1} & \Bf 28.0 & \Bf 50.9 \\
        \addlinespace[0.6ex]
        \cdashline{2-12}
        \addlinespace[0.6ex]
        & \multirow{2}{*}{\antelope{}-TG} & Transformer & 19.4 & 40.5 & 51.3 & \Uline{71.6} & 67.1 & 33.8 & 55.9 & 24.4 & 49.2 \\
        & & Gated DeltaNet & 18.1 & 40.4 & \Uline{51.5} & 71.3 & 68.6 & 34.8 & \Bf 57.5 & 25.6 & 50.0 \\

        \midrule
        \multirow{7}{*}{\rotatebox[origin=c]{90}{\textbf{810M}}}
        & \multirow{3}{*}{Baseline} & Transformer & 13.6 & 46.2 & 57.0 & 73.1 & 71.3 & 37.3 & 58.3 & 28.6 & 53.1 \\
        & & Mamba-2 & 13.4 & 45.6 & 58.2 & 72.7 & 72.8 & \Bf 40.1 & 56.0 & \Bf 31.2 & 53.8 \\
        & & Gated DeltaNet & \Uline{12.1} & 47.4 & \Uline{58.3} & 72.5 & 73.3 & \Uline{39.8} & 58.6 & 29.6 & 54.2 \\
        \addlinespace[0.6ex]
        \cdashline{2-12}
        \addlinespace[0.6ex]
        & \multirow{2}{*}{\antelope{}-TM} & Transformer & 12.5 & \Uline{48.0} & 58.0 & \Uline{73.6} & \Bf 73.7 & 39.4 & 59.1 & \Uline{31.0} & \Bf 54.7 \\
        & & Mamba-2 & \Bf 11.9 & \Bf 48.2 & 57.9 & \Bf 73.9 & \Bf 73.7 & 39.0 & \Bf 59.6 & 30.6 & \Bf 54.7 \\
        \addlinespace[0.6ex]
        \cdashline{2-12}
        \addlinespace[0.6ex]
        & \multirow{2}{*}{\antelope{}-TG} & Transformer & 13.4 & 46.2 & 58.1 & 72.8 & 71.5 & 37.5 & 58.5 & 27.8 & 53.2 \\
        & & Gated DeltaNet & 12.8 & 45.8 & \Bf 58.4 & 73.1 & 72.4 & 38.5 & \Uline{59.4} & 30.8 & 54.0 \\
        \midrule

        \multirow{7}{*}{\rotatebox[origin=c]{90}{\textbf{1.4B}}}
        & \multirow{3}{*}{Baseline} & Transformer & 11.4 & 49.9 & 60.6 & 74.1 & 73.6 & 42.1 & 58.0 & 31.2 & 55.6 \\
        & & Mamba-2 & 11.2 & 48.6 & 60.9 & 74.7 & 74.5 & 42.7 & 58.1 & 30.4 & 55.7 \\
        & & Gated DeltaNet & \Uline{10.6} & 49.9 & 61.9 & 75.0 & 75.0 & 42.2 & \Uline{60.9} & 31.4 & \Uline{56.6} \\
        \addlinespace[0.6ex]
        \cdashline{2-12}
        \addlinespace[0.6ex]
        & \multirow{2}{*}{\antelope{}-TM} & Transformer & 11.1 & \Uline{50.2} & 61.3 & 75.0 & \Uline{75.7} & 42.0 & 58.0 & 31.6 & 56.3 \\
        & & Mamba-2 & \Bf 10.5 & \Bf 50.4 & \Bf 62.1 & \Bf 75.2 & 75.3 & \Bf 43.6 & 58.5 & 31.2 & \Uline{56.6} \\
        \addlinespace[0.6ex]
        \cdashline{2-13}
        \addlinespace[0.6ex]
        & \multirow{2}{*}{\antelope{}-TG} & Transformer & 10.9 & 49.9 & 61.8 & 74.9 & 75.4 & 41.7 & 59.8 & \Uline{31.8} & 56.5 \\
        & & Gated DeltaNet & \Uline{10.6} & 50.0 & \Bf 62.1 & \Uline{75.1} & \Bf 76.2 & \Uline{43.1} & \Bf 62.0 & \Bf 32.2 & \Bf 57.3 \\
        \bottomrule
    \end{tabular}
    }
    \label{tab:downstream_evaluation}
\end{table}
\section{Empirical Results and Properties of~\antelope}
\label{sec:results}
We evaluate the empirical performance and capabilities of \antelope{} in the following section.
\Cref{sec:results:subsec:isolated} discusses the language modeling (\Cref{sec:results:subsec:lm}) and retrieval (\Cref{sec:results:subsec:retrieval}) performance of our models when operating using only one mixer mode.
\Cref{sec:results:subsec:mode-switch} then explores the mode switching capabilities of the model, and \Cref{sec:results:subsec:ablations} ablates architectural and training design choices.

\paragraph{\antelope{} Setup.}
Our model follows the Transformer++ setup originating from~\citet{touvron2023llamaopenefficientfoundation} and detailed in~\citet{gu2024mambalineartimesequencemodeling}, which includes interleaved SwiGLU MLPs.
We use \antelope-TM to denote a self-attention (Transformer)/Mamba-2 shared-block model, and \antelope-TG to denote a self-attention/GDN shared-block model.
We set the dimensions of the query, key, and value to be equal, i.e., $D_k=D_v$ following standard Transformer convention; however, we fix the head dimension $D_{k,v}=128$ across all scales to preserve the recurrent state size.
As the gate increases the active parameters, the resulting mixer-to-MLP parameter ratio increases from the baselines Transformer's 4:8 to 5:8.
As the MLPs and most sequence-mixer projections are shared across mixers, more than $90\%$ of the weights are jointly used across modes.\footnote{\,The percentage calculation excludes the embedding in the total parameter count.}

\paragraph{Baseline Setup.}
Transformer baselines follow the Transformer++ architecture of~\citet{touvron2023llamaopenefficientfoundation} and head structure of~\citet{brown2020languagemodelsfewshotlearners}.
Mamba-2 baselines interleave Mamba-2 blocks, using $D_k=128,D_v=64$ and expansion factor of 2, with SwiGLU MLPs at a 6:6 parameter ratio.
Gated DeltaNet baselines similarly interleave GDN blocks, using $D_k=128, D_v=256$, with MLPs at the same 6:6 ratio.
To parameter-match models, we increase the MLP widths of the baselines to compensate for the gate in \antelope{} blocks (the short convolution leads to a negligible increase).

\paragraph{Experimental Setup.}
We pretrained models at four different scales, each with 100B tokens of FineWeb-Edu~\citep{lozhkov2024fineweb-edu} using the GPT-2 tokenizer~\citep{Radford2019LanguageMA} at 2K context length.
A cosine scheduler was used with $10\%$ of total steps allocated to warmup, and AdamW~\citep{loshchilov2019decoupledweightdecayregularization} with $\beta=(0.9,0.95)$ and $0.1$ weight decay was used as the optimizer.
For baselines, peak learning rate was set to $5\times$ that of~\citet{brown2020languagemodelsfewshotlearners}, following \citet{dao2024transformersssmsgeneralizedmodels}.
A $10\times$ factor was used for~\antelope{} models.
We find the increase improves both mixer performances while preserving training stability, potentially mitigating reduced mixer-specific gradients from chunked-training and different optimization landscapes induced by tied representations.
Training used bfloat16 mixed precision and a global batch size of 1M tokens for 1B models, and 0.5M for the rest.

\paragraph{Evaluation Setup.}
The language modeling abilities of models were evaluated with a suite of standard commonsense reasoning and language understanding tasks: LAMBADA~\citep{paperno2016lambadadatasetwordprediction,Radford2019LanguageMA}, HellaSwag~\citep{zellers2019hellaswagmachinereallyfinish}, PIQA~\citep{bisk2019piqareasoningphysicalcommonsense}, Arc-Easy and Challenge~\citep{clark2018thinksolvedquestionanswering}, WinoGrande~\citep{sakaguchi2019winograndeadversarialwinogradschema}, and OpenbookQA~\citep{mihaylov2018suitarmorconductelectricity}.
We also measured the retrieval capabilities of the 1.4B models on both synthetic needle-in-the-haystack (NIAH) tasks~\citep{hsieh2024rulerwhatsrealcontext} and real-world retrieval tasks in cloze format~\citep{arora2024justreadtwiceclosing,arora2025simplelinearattentionlanguage}:
SWDE~\citep{arora2025languagemodelsenablesimple,lockard-etal-2019-openceres},
SQuAD~\citep{rajpurkar2018know},
FDA~\citep{arora2025languagemodelsenablesimple},
TriviaQA (TQA)~\citep{joshi2017triviaqalargescaledistantly},
NQ~\citep{kwiatkowski-etal-2019-natural}, and
DROP~\citep{dua2019dropreadingcomprehensionbenchmark}.
A context length of 2K was used for all retrieval tasks.

\subsection{Language Modeling and Retrieval with Individual Mixer Modes}
\label{sec:results:subsec:isolated}
\subsubsection{Language Modeling Performance of Individual Mixers}
\label{sec:results:subsec:lm}
\Cref{tab:downstream_evaluation} reports each \antelope{} model's performance when using one of its mixer modes in isolation, i.e., either softmax attention or linear mechanism used for the entire sequence, to determine whether each mode remains useful on its own.
Across scales, \antelope{} modes remain competitive with their corresponding single-mixer baselines \emph{despite} being trained for the same number of tokens.
At the 1.4B scale, both attention and linear modes of \antelope-TG and \antelope-TM outperform the baselines by at least 0.7 percentage points on average.
Notably, despite these tasks being evaluated ``out-of-distribution'' --- the mode switching training does not explicitly train the entire model with all chunks assigned to the same mixer --- the models are still able to generalize to this edge case (the probability that a sample is processed entirely by the linear mechanism is $(3/4)^{16}\approx 1\%$ and $(1/4)^{16}\approx 2.3\times 10^{-8}\%$ for softmax attention).
These results suggest that sharing key-value representations and tying the majority of weights do \emph{not} prevent either mixer from learning effective standalone behavior.

\subsubsection{Retrieval Capabilities of Individual Mixers}
\label{sec:results:subsec:retrieval}
 
\begin{table*}[!t]
    \small
    \centering
    \caption{
    Retrieval results for 1.4B baseline and \antelope{} models on real-world and synthetic retrieval tasks at 2K context length.
    Results reported below use the same mixer for the entire model and sequence.
    \antelope{}-TM denotes Transformer/Mamba-2 shared blocks, and \antelope{}-TG, Transformer/Gated DeltaNet.
    The isolated modes of \antelope{} remain comparable in to their corresponding baselines.
    }
    \resizebox{\textwidth}{!}{
    \sisetup{
        detect-weight,
        table-align-text-post=false,
        table-space-text-post={*}
    }
    \begin{tabular}{l l *{11}{S[table-format=2.1]}}
        \toprule
        \multicolumn{2}{c}{}
        & \multicolumn{7}{c}{Real-World Retrieval}
        & \multicolumn{4}{c}{Synthetic Retrieval} \\
        \cmidrule(lr){3-9} \cmidrule(lr){10-13}
        \textbf{Family} & \textbf{Mixer}
        & {SWDE} & {SQuAD} & {FDA} & {TQA} & {NQ} & {DROP} & {\textbf{Avg.}}
        & {NIAH-1} & {NIAH-2} & {NIAH-3} & {\textbf{Avg.}} \\
        \midrule

        Baseline & \multirow{3}{*}{Transformer}
        & 42.3 & 42.8 & 55.2 & 65.9 & 26.7 & 22.9 & 42.6
        & 99.0 & 90.4 & 95.0 & 94.8 \\

        \antelope{}-TM & 
        & 50.1 & 41.7 & 57.6 & 63.9 & 28.3 & 22.5 & 44.0
        & 99.8 & 99.4 & 81.6 & 93.6 \\

        \antelope{}-TG & 
        & 44.5 & 41.2 & 46.4 & 64.8 & 27.8 & 21.7 & 41.1
        & 99.4 & 98.6 & 99.4 & 99.1 \\

        \midrule

        Baseline & \multirow{2}{*}{Mamba-2}
        & 20.6 & 36.1 & 21.0 & 61.7 & 22.4 & 20.3 & 30.3
        & 90.4 & 15.8 & 33.2 & 46.5 \\

        \antelope{}-TM &
        & 22.2 & 36.4 & 19.4 & 63.7 & 21.8 & 22.3 & 31.0
        & 50.8 & 17.0 & 41.0 & 36.3 \\

        \midrule

        Baseline & \multirow{2}{*}{Gated DeltaNet}
        & 27.4 & 36.4 & 24.2 & 63.3 & 23.8 & 21.6 & 32.8
        & 99.4 & 39.6 & 32.6 & 57.2 \\

        \antelope{}-TG &
        & 29.3 & 37.9 & 18.2 & 64.0 & 23.5 & 22.0 & 32.5
        & 99.4 & 60.2 & 82.8 & 80.8 \\

        \bottomrule
    \end{tabular}
    }
    \label{tab:retrieval_single_mode}
\end{table*}

The single-mode retrieval results show that \antelope{} generally preserves the retrieval capabilities of its constituent mixers when using the same mixer for the entire sequence.
Both Mamba-2 and GDN variants achieve comparable real-world retrieval performance except for the GDN variant on the FDA dataset, which requires extracting information from unstructured data.
Notably, the GDN mode of \antelope{}-TG \emph{substantially} improves NIAH performance over its baseline, achieving more than $2\times$ the accuracy on NIAH-3 and more than $1.5\times$ the accuracy on NIAH-2.
We emphasize that these results on the linear mechanisms were achieved despite using only half and two-thirds of the total state size of the respective Mamba-2 and GDN baselines, which suggests that shared-block model and training can improve downstream capabilities.\footnote{\,The total recurrent state size of Mamba-2 is $2D_kD_\text{model}$, GDN is $1.5D_kD_\text{model}$, and \antelope{} is $D_kD_\text{model}$.}

\subsection{Flexible Mode Switching during Inference}
\label{sec:results:subsec:mode-switch}
\begin{figure}[!t]
    \centering
    \includegraphics[width=\textwidth]{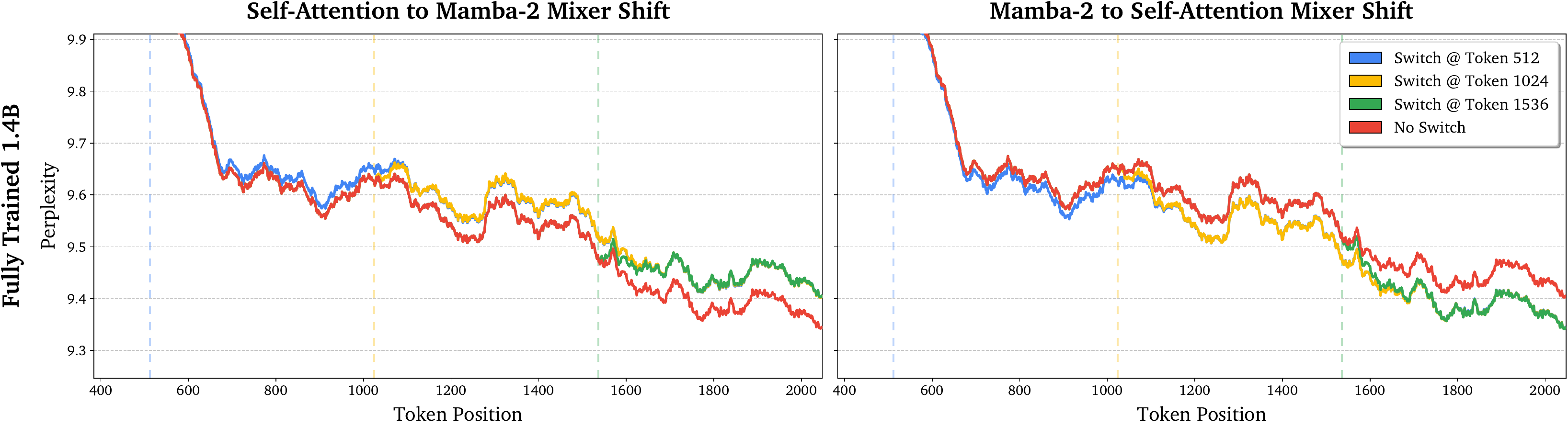}
    \caption{
    Smoothed token-level perplexity across token position for pretrained 1.4B \antelope{}-TM model trained with chunked mixed-mode training.
    After switching from softmax attention to Mamba-2 and vice versa at different positions, perplexity rapidly approaches the corresponding no-switch baseline, indicating that the mixers share compatible representations.
    }
    \label{fig:mode-switch-1B} %
\end{figure}

While~\antelope{} can be deployed in an attention-only or linear-only mode, our chunk-level mode switching mechanism enables a new axis of exploring hybrid models.
The models can flexibly change between mixers during prefill with little to no degradation in perplexity as demonstrated in~\Cref{fig:mode-switch-1B}.
In the pretrained \antelope{}-TM 1.4B model, after a switch, the perplexity rapidly approaches that of the corresponding no-switch baseline in the selected mode.
The results in \Cref{sec:additional-mode-switching} display the same behaviors for non-chunk-aligned switch boundaries and multiple switches for both TM and TG models, demonstrating the ability of various mixers to share internal representations across sequences.
\begin{table*}[!t]
    \small
    \centering
    \caption{
    Cross-mode retrieval results for 1.4B baseline and \antelope{} models on real-world and synthetic retrieval tasks at 2K context length, grouped by the mixer used for prompt prefill and generation (\textbf{Prompt + Gen}).
    Baselines use the same mixer for the entire sequence, while \antelope{} use a different mixer for prefilling the context (\textbf{Context}) than for the prompt prefill and generation.
    The \antelope{} modes can remain comparable to baselines despite switching modes, suggesting that the underlying representations required for retrieval are preserved across modes.
    }
    \resizebox{\textwidth}{!}{
    \sisetup{
        detect-weight,
        table-align-text-post=false,
        table-space-text-post={*}
    }
    \begin{tabular}{l l l *{11}{S[table-format=2.1]}}
        \toprule
        \multicolumn{3}{c}{}
        & \multicolumn{7}{c}{Real-World Retrieval}
        & \multicolumn{4}{c}{Synthetic Retrieval} \\
        \cmidrule(lr){4-10} \cmidrule(lr){11-14}
        \textbf{Prompt + Gen} & \textbf{Family} & \textbf{Context}
        & {SWDE} & {SQuAD} & {FDA} & {TQA} & {NQ} & {DROP} & {\textbf{Avg.}}
        & {NIAH-1} & {NIAH-2} & {NIAH-3} & {\textbf{Avg.}} \\
        \midrule

        \multirow{3}{*}{Transformer}
        & Baseline & Transformer
        & 42.3 & 42.8 & 55.2 & 65.9 & 26.7 & 22.9 & 42.6
        & 99.0 & 90.4 & 95.0 & 94.8 \\

        & \antelope{}-TM & Mamba-2
        & 50.9 & 40.9 & 56.4 & 64.5 & 28.2 & 22.4 & 43.9
        & 99.8 & 97.6 & 58.0 & 85.1 \\

        & \antelope{}-TG & Gated DeltaNet
        & 46.1 & 41.1 & 47.9 & 64.1 & 28.0 & 20.9 & 41.4
        & 99.0 & 97.6 & 95.8 & 97.5 \\

        \midrule

        \multirow{2}{*}{Mamba-2}
        & Baseline & Mamba-2
        & 20.6 & 36.1 & 21.0 & 61.7 & 22.4 & 20.3 & 30.4
        & 90.4 & 15.8 & 33.2 & 46.5 \\

        & \antelope{}-TM & Transformer
        & 21.8 & 37.1 & 17.8 & 63.2 & 21.3 & 21.6 & 30.5
        & 44.6 & 14.6 & 51.0 & 36.7 \\

        \midrule

        \multirow{2}{*}{Gated DeltaNet}
        & Baseline & Gated DeltaNet
        & 27.4 & 36.4 & 24.2 & 63.3 & 23.8 & 21.6 & 32.8
        & 99.4 & 39.6 & 32.6 & 57.2 \\

        & \antelope{}-TG & Transformer
        & 29.5 & 38.2 & 17.1 & 63.9 & 24.0 & 21.2 & 32.3
        & 99.4 & 73.6 & 88.2 & 87.1 \\

        \bottomrule
    \end{tabular}
    }
    \label{tab:retrieval_cross_mode}
\end{table*}

We further evaluate the mode switching ability on retrieval tasks by splitting each sample into a context segment (\textbf{Context}) and a prompt/generation segment (\textbf{Prompt + Gen}), then processing the two segments with different modes.
For synthetic NIAH tasks, the context consists of the haystack, and the prompt consists of the needle query.
For real-world tasks, because the boundary between context and prompt is less distinct due to the cloze format, we designate the first 97.5\% of tokens as the context and the remaining tokens as the prompt.
\Cref{tab:retrieval_cross_mode} shows that mode-switching often preserves the retrieval performance of the target prompt/generation mixer, especially on real-world retrieval tasks.
We highlight that both \antelope{} models are able to prefill with the linear mode and generate with softmax attention, achieving comparable results to the Transformer baseline and \emph{significantly surpassing their linear baselines}.
In particular, with mixed inference mode, \antelope{}-TM achieves 13.5 percentage points higher on average for real-world retrieval tasks, and 38.6 percentage points higher for NIAH tests.
For \antelope{}-TG, the numbers are 8.6 and 40.3, respectively.
We note that synthetic evaluations are more sensitive to switching direction and mixer choice, particularly the Transformer-to-Mamba-2 setting.
These results suggest that exact retrieval may stress the compatibility of shared representations more strongly than real-world retrieval for certain model configurations.

This new capability may enable use cases such as models that vary mechanisms depending on the task, e.g., softmax attention for retrieval-heavy questions, or paradigms where large portions of thinking traces are generated with the linear mixer and verified with the attention one.
We note that models that deploy with mode switching enabled will require the storage of both the KV cache and linear RNN state, but memory costs would be dominated by that of the KV cache at longer context lengths.

\subsection{Architecture and Training Ablations}
\label{sec:results:subsec:ablations}
In the following section, we ablate the architectural and training choices that enable the strong performance and mode switching abilities of \antelope{} models.
We explore the impact of the mixer-specific query and additional linear model components on performance and the importance of the chunked mixed-mode training on mode-switching capabilities.
Ablations are conducted at the 350M scale, unless specified otherwise, with the standard training regime as the final models except trained to Chinchilla scaling law token count ($20\times$ tokens-to-parameter ratio)~\citep{hoffmann2022trainingcomputeoptimallargelanguage}.
When ablating our \antelope{} architecture, we utilize sequence mixed-mode training, where an entire sequence is assigned to only one mixer instead of our final chunked mixed-mode training, as both training regimes result in comparable pretraining loss and findings are consistent under both.

\paragraph{Mixer-Specific Queries.}
The final shared \antelope{} block uses mixer-specific query projections while sharing key and value projections across mixer modes.
While query weights can be shared across mixers, which would reduce the total parameter count, it would not reduce the active parameter count and empirically degrades performance (\Cref{tab:query_ablation}).
Thus, we can conclude that attention and recurrent mixers can share the state updating representations, i.e., the key and values, but benefit from separate readout components, i.e., the query, due to their differences in state parameterization.
We find similar mechanism-specific design requirements for \antelope{}-TG.
For instance, SiLU activations after the short convolution are important, and the query-key normalization should only be applied to the GDN components (\Cref{sec:additional-ablations}).

\paragraph{Additional Architectural Components.}
Our ablations find that the usage of the short convolution is critical for both softmax attention and the linear model performance, while adding the gate further improves perplexity (\Cref{tab:conv_gate_ablation}).
In contrast, applying element-wise gating without the short convolution hurts both performances, underscoring the complex interactions between components that arise when designing shared blocks.
Because \antelope{} uses separate queries for each mixer and supports mode switching during inference, we apply the short convolution only to the shared keys and values.
Notably, when these components, normalization, convolution, and gate, are added to the Transformer baseline, the language modeling performance does not improve (\Cref{tab:baseline_abl}).
Thus, the strong \antelope{} results observed are unlikely to be caused solely by the addition of these components, but rather by the interaction of the various inductive biases when sharing core representations.

\begin{table}[!t]
    \small
    \centering
    \caption{
    \textbf{Left:} Untied, or disjoint, query weights outperform shared query weights for both types of RMS normalization, and default RMSNorm outperforms grouped RMSNorm.
    \textbf{Right:} Adding both an element-wise gate and short convolution on the key and values of a disjoint query model leads to the best performance.
    Model parameter counts are matched across variants; gated models increase the parameter count in the mixer block, so the MLP width is increased in non-gated models to compensate.
    }
    \resizebox{\textwidth}{!}{
        \sisetup{
            detect-weight,
            table-align-text-post=false,
            table-space-text-post={*},
            table-column-width=2.2cm 
        }
        
        \begin{tabular}{l l S S}
            \toprule
            \textbf{Query} & \textbf{RMSNorm} & {Attn ppl $\downarrow$\;} & {Mamba-2 ppl $\downarrow$} \\
            \midrule
            \multirow{2}{*}{\shortstack[l]{Shared}} 
            & Default & 16.10 & 16.00 \\
            & Grouped & 17.73 & 17.77 \\
            \midrule
            \multirow{2}{*}{\shortstack[l]{Disjoint}} 
            & Default & 15.93 & 15.82 \\
            & Grouped & 17.47 & 17.57 \\
            \bottomrule
        \end{tabular}
        \phantomsubcaption\label{tab:query_ablation}

        \hspace{0.25cm} 
        
        \begin{tabular}{c c S S}
            \toprule
            \textbf{Conv?} & \textbf{Gate?} & {Attn ppl $\downarrow$\;} & {Mamba-2 ppl $\downarrow$} \\
            \midrule
            \multirow{2}{*}{\shortstack[l]{\cmark}} 
            & \cmark & 15.09 & 15.28 \\
            & \xmark & 15.35 & 15.36 \\
            \midrule
            \multirow{2}{*}{\shortstack[l]{\xmark}} 
            & \cmark & 16.67 & 16.59 \\
            & \xmark & 15.91 & 15.78 \\
            \bottomrule
        \end{tabular}
        \phantomsubcaption\label{tab:conv_gate_ablation}
    }
\end{table}

\begin{table}[!t]
    \small
    \centering
    \caption{
    \textbf{Left:} Additional architecture components added to fully pretrained, parameter-matched Transformer baselines do not benefit downstream language modeling performance.
    \textbf{Right:} The performance of chunked mixed-mode training is comparable to sequence mixed-mode training but results in stronger, more robust mode switching abilities, highlighted in \Cref{fig:mode-switch-1B} and \Cref{fig:mode-switch-abl}.
    }
    \resizebox{\textwidth}{!}{
        \sisetup{
            detect-weight,
            table-align-text-post=false,
            table-space-text-post={*},
            table-column-width=2.2cm 
        }
        
        \begin{tabular}{l S S}
            \toprule
            \textbf{Transformer 1.4B} & {Avg LM acc $\uparrow$} \\
            \midrule
            Baseline & 55.6 \\
            + Norm & 55.4 \\
            + Norm + Conv + Gate & 55.6 \\
            \bottomrule
        \end{tabular}
        \phantomsubcaption\label{tab:baseline_abl}

        \hspace{0.5cm} 

        \begin{tabular}{l S S}
            \toprule
            \textbf{Training} & {Attn ppl $\downarrow$} & {Mamba-2 ppl $\downarrow$} \\
            \midrule
            Sequence Assignment & 15.09 & 15.28 \\
            \midrule
            Chunk Assignment & 15.08 & 15.27 \\
            \bottomrule
        \end{tabular}
        \phantomsubcaption\label{tab:training_ablation}
    }
\end{table}

\begin{figure}[!t]
    \centering
    \includegraphics[width=\textwidth]{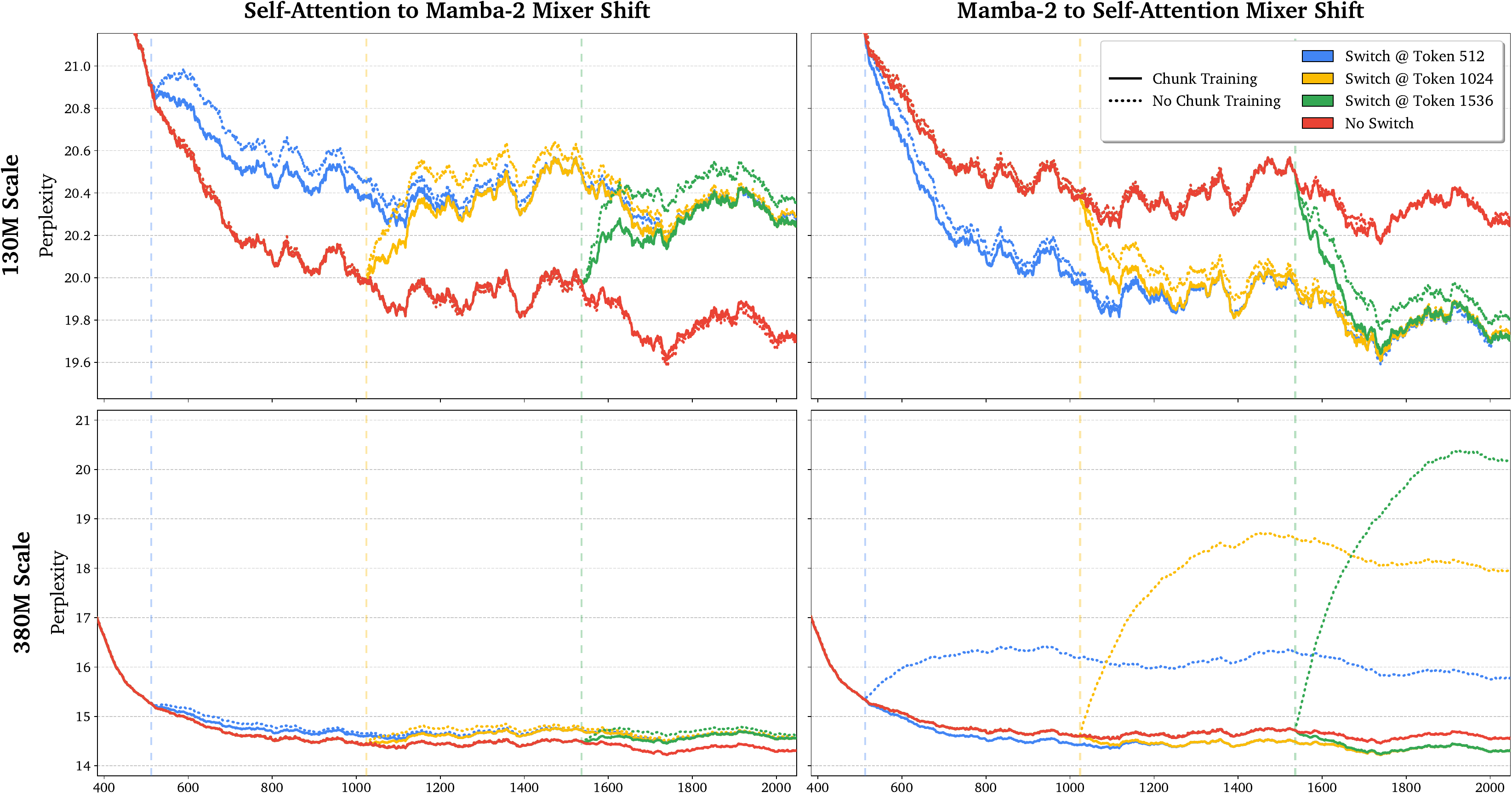}
    \caption{Smoothed perplexity across token index position for \antelope{}-TM models trained with and without chunk-level switching. Models trained without mode chunk switching perform slightly worse after switching mixers and can suffer from strong degradation in some cases.
    }
    \label{fig:mode-switch-abl}
\end{figure}

\paragraph{Chunked Mixed-Mode Training.}
Our chunk-level mixed-mode training regime is critical in enabling robust mode switching for our \antelope{} model.
As an alternative, we train our models with a sequence-level mixed-mode scheme where an entire training sequence is processed by only one mixer.
While these models achieve comparable pretraining losses to chunk-level (\Cref{tab:training_ablation}), the mode switching capability does \emph{not consistently} manifest.
The sequence-level trained models seem to possess mode switching capabilities from softmax attention to Mamba-2, but suffer from massive perplexity degradation when shifting from Mamba-2 to softmax attention at certain scales.
\Cref{fig:mode-switch-abl} displays the smoothed perplexities at each token position for the 125M and 350M models trained with and without chunk switching, and \Cref{sec:additional-mode-switching} discusses further.
Our ablation suggests that the mixer transitions induced by the chunk-based training encourages the representations to remain more compatible throughout a sequence as compared to sequence-based training.

We note that chunked mixed-mode training incurs an increase in computational overhead compared to sequence-level mixed training.
While softmax attention strictly forgoes the quadratic cost of calculating outputs for unselected sequence chunks, the linear RNN must roll its recurrent state forward across all chunks to preserve output correctness, resulting in a fixed linear compute cost.
We analyze general compute usage in \Cref{sec:compute-analysis} and leave overhead reduction for future work.

\section{Related Work}
While some methodologies have been introduced that can change the computational pathway for a given input, e.g., context-dependent segmentation for tokenization~\citep{hwang2025dynamicchunkingendtoendhierarchical}, there is comparatively little prior work that is directly motivated by the prospect of using different sequence mixers throughout generation or prefill.
Our method, to the best of our knowledge, is among the first to study the flexible switching of sequence mixers in an autoregressive language modeling setting.
Concurrent work HAM~\citep{lufkin2026hybridassociativememories} combines a linear RNN with attention by using the linear model to compress the full sequence while reserving the KV cache for certain routed tokens, resulting in linear RNN and sparse attention hybrid.
TransMamba~\citep{li2025transmambaflexiblyswitchingtransformer} processes a sequence with a mixture of softmax attention and Mamba-2 with shared weights, but has a predetermined, fixed switch location at each layer.
This constraint prevents the flexible usage of various mixers depending on the sequence context, which is further compounded by their unidirectional switch, i.e., only from softmax attention to Mamba once.
Recent SLA2~\citep{zhang2026sla2sparselinearattentionlearnable} also enables mixer-choice through a learned router but is formulated as sparse-linear attention for video diffusion rather than flexible switching for autoregressive prefill or generation.
Our work provides a general framework for mixer selection in autoregressive language modeling and can serve as a basis for objectives such as learned routing or learned sparsity.
We further cover related works in \Cref{sec:additional-related}.

\section{Discussion and Conclusion}
In our work, we introduce \antelope{}, a multi-mixer architecture that enables sequence-axis hybridization where different mixer mechanisms can be utilized across a sequence.
\antelope{} maintains shared representations across the softmax attention and linear mechanism through tied projections, and achieves strong language modeling performance, preserves retrieval capabilities, and supports mode switching during prefill and generation. 
One consideration to highlight is that our design requires the model to keep and update both the KV cache and state memory at each timestep throughout generation if mode switching is used, but for longer context, the memory cost is dominated by the KV cache.
Reducing the overhead of maintaining multiple states remains an important direction.

Our results suggest several extensions for developing sequence-level hybrid models.
While our current work relies on static assignment, there exist many approaches to learnable mode-switching, e.g., using RL to train routers that dynamically route tokens using FLOPs saved as the reward.
In addition, because the \antelope{} design can naturally incorporate other sequence mixers, certain layers that share similar head structures to softmax attention or can utilize its KV cache, e.g., 2-simplicial attention~\citep{roy2025fastsimplex2simplicialattention}, may further improve performance.
Finally, the training dynamics of our multi-mixer models remain largely unexplored and may be vastly different than that of standard models.
As the number of mixers or model scale increases, it remains to be seen whether techniques, such as discriminative learning rates or specialized training objectives, may become more important. 

More broadly, the sequence-axis hybrid paradigm \emph{opens a new design space} for adaptive compute.
For instance, linear recurrent modes could serve as efficient drafters for speculative decoding or could rapidly generate the longer intermediate reasoning traces prior to fully synthesizing the answer with quadratic attention.
Beyond these applications, our work suggests that the representations learned by different mechanisms can overlap and be used jointly, raising crucial questions of when and how different models develop compatible understandings and representations.

\bibliography{main}

@misc{dao2024transformersssmsgeneralizedmodels,
      title={Transformers are SSMs: Generalized Models and Efficient Algorithms Through Structured State Space Duality}, 
      author={Tri Dao and Albert Gu},
      year={2024},
      eprint={2405.21060},
      archivePrefix={arXiv},
      primaryClass={cs.LG},
      url={https://arxiv.org/abs/2405.21060}, 
}

@misc{katharopoulos2020transformersrnnsfastautoregressive,
      title={Transformers are RNNs: Fast Autoregressive Transformers with Linear Attention}, 
      author={Angelos Katharopoulos and Apoorv Vyas and Nikolaos Pappas and François Fleuret},
      year={2020},
      eprint={2006.16236},
      archivePrefix={arXiv},
      primaryClass={cs.LG},
      url={https://arxiv.org/abs/2006.16236}, 
}

@misc{gu2024mambalineartimesequencemodeling,
      title={Mamba: Linear-Time Sequence Modeling with Selective State Spaces}, 
      author={Albert Gu and Tri Dao},
      year={2024},
      eprint={2312.00752},
      archivePrefix={arXiv},
      primaryClass={cs.LG},
      url={https://arxiv.org/abs/2312.00752}, 
}

@misc{hua2022transformerqualitylineartime,
      title={Transformer Quality in Linear Time}, 
      author={Weizhe Hua and Zihang Dai and Hanxiao Liu and Quoc V. Le},
      year={2022},
      eprint={2202.10447},
      archivePrefix={arXiv},
      primaryClass={cs.LG},
      url={https://arxiv.org/abs/2202.10447}, 
}

@misc{du2025momlinearsequencemodeling,
      title={MoM: Linear Sequence Modeling with Mixture-of-Memories}, 
      author={Jusen Du and Weigao Sun and Disen Lan and Jiaxi Hu and Yu Cheng},
      year={2025},
      eprint={2502.13685},
      archivePrefix={arXiv},
      primaryClass={cs.CL},
      url={https://arxiv.org/abs/2502.13685}, 
}

@misc{dong2024hymbahybridheadarchitecturesmall,
      title={Hymba: A Hybrid-head Architecture for Small Language Models}, 
      author={Xin Dong and Yonggan Fu and Shizhe Diao and Wonmin Byeon and Zijia Chen and Ameya Sunil Mahabaleshwarkar and Shih-Yang Liu and Matthijs Van Keirsbilck and Min-Hung Chen and Yoshi Suhara and Yingyan Lin and Jan Kautz and Pavlo Molchanov},
      year={2024},
      eprint={2411.13676},
      archivePrefix={arXiv},
      primaryClass={cs.CL},
      url={https://arxiv.org/abs/2411.13676}, 
}

@misc{wu2018groupnormalization,
      title={Group Normalization}, 
      author={Yuxin Wu and Kaiming He},
      year={2018},
      eprint={1803.08494},
      archivePrefix={arXiv},
      primaryClass={cs.CV},
      url={https://arxiv.org/abs/1803.08494}, 
}

@misc{waleffe2024empiricalstudymambabasedlanguage,
      title={An Empirical Study of Mamba-based Language Models}, 
      author={Roger Waleffe and Wonmin Byeon and Duncan Riach and Brandon Norick and Vijay Korthikanti and Tri Dao and Albert Gu and Ali Hatamizadeh and Sudhakar Singh and Deepak Narayanan and Garvit Kulshreshtha and Vartika Singh and Jared Casper and Jan Kautz and Mohammad Shoeybi and Bryan Catanzaro},
      year={2024},
      eprint={2406.07887},
      archivePrefix={arXiv},
      primaryClass={cs.LG},
      url={https://arxiv.org/abs/2406.07887}, 
}

@misc{arora2025simplelinearattentionlanguage,
      title={Simple linear attention language models balance the recall-throughput tradeoff}, 
      author={Simran Arora and Sabri Eyuboglu and Michael Zhang and Aman Timalsina and Silas Alberti and Dylan Zinsley and James Zou and Atri Rudra and Christopher Ré},
      year={2025},
      eprint={2402.18668},
      archivePrefix={arXiv},
      primaryClass={cs.CL},
      url={https://arxiv.org/abs/2402.18668}, 
}

@misc{kimiteam2025kimilinearexpressiveefficient,
      title={Kimi Linear: An Expressive, Efficient Attention Architecture}, 
      author={{Kimi Team} and Yu Zhang and Zongyu Lin and Xingcheng Yao and Jiaxi Hu and Fanqing Meng and Chengyin Liu and Xin Men and Songlin Yang and Zhiyuan Li and Wentao Li and Enzhe Lu and Weizhou Liu and Yanru Chen and Weixin Xu and Longhui Yu and Yejie Wang and Yu Fan and Longguang Zhong and Enming Yuan and Dehao Zhang and Yizhi Zhang and T. Y. Liu and Haiming Wang and Shengjun Fang and Weiran He and Shaowei Liu and Yiwei Li and Jianlin Su and Jiezhong Qiu and Bo Pang and Junjie Yan and Zhejun Jiang and Weixiao Huang and Bohong Yin and Jiacheng You and Chu Wei and Zhengtao Wang and Chao Hong and Yutian Chen and Guanduo Chen and Yucheng Wang and Huabin Zheng and Feng Wang and Yibo Liu and Mengnan Dong and Zheng Zhang and Siyuan Pan and Wenhao Wu and Yuhao Wu and Longyu Guan and Jiawen Tao and Guohong Fu and Xinran Xu and Yuzhi Wang and Guokun Lai and Yuxin Wu and Xinyu Zhou and Zhilin Yang and Yulun Du},
      year={2025},
      eprint={2510.26692},
      archivePrefix={arXiv},
      primaryClass={cs.CL},
      url={https://arxiv.org/abs/2510.26692}, 
}

@misc{nvidia2025nvidianemotron3efficient,
      title={NVIDIA Nemotron 3: Efficient and Open Intelligence}, 
      author={NVIDIA and : and Aaron Blakeman and Aaron Grattafiori and Aarti Basant and Abhibha Gupta and Abhinav Khattar and Adi Renduchintala and Aditya Vavre and Akanksha Shukla and Akhiad Bercovich and Aleksander Ficek and Aleksandr Shaposhnikov and Alex Kondratenko and Alexander Bukharin and Alexandre Milesi and Ali Taghibakhshi and Alisa Liu and Amelia Barton and Ameya Sunil Mahabaleshwarkar and Amir Klein and Amit Zuker and Amnon Geifman and Amy Shen and Anahita Bhiwandiwalla and Andrew Tao and Anjulie Agrusa and Ankur Verma and Ann Guan and Anubhav Mandarwal and Arham Mehta and Ashwath Aithal and Ashwin Poojary and Asif Ahamed and Asit Mishra and Asma Kuriparambil Thekkumpate and Ayush Dattagupta and Banghua Zhu and Bardiya Sadeghi and Barnaby Simkin and Ben Lanir and Benedikt Schifferer and Besmira Nushi and Bilal Kartal and Bita Darvish Rouhani and Boris Ginsburg and Brandon Norick and Brandon Soubasis and Branislav Kisacanin and Brian Yu and Bryan Catanzaro and Carlo del Mundo and Chantal Hwang and Charles Wang and Cheng-Ping Hsieh and Chenghao Zhang and Chenhan Yu and Chetan Mungekar and Chintan Patel and Chris Alexiuk and Christopher Parisien and Collin Neale and Cyril Meurillon and Damon Mosk-Aoyama and Dan Su and Dane Corneil and Daniel Afrimi and Daniel Lo and Daniel Rohrer and Daniel Serebrenik and Daria Gitman and Daria Levy and Darko Stosic and David Mosallanezhad and Deepak Narayanan and Dhruv Nathawani and Dima Rekesh and Dina Yared and Divyanshu Kakwani and Dong Ahn and Duncan Riach and Dusan Stosic and Edgar Minasyan and Edward Lin and Eileen Long and Eileen Peters Long and Elad Segal and Elena Lantz and Ellie Evans and Elliott Ning and Eric Chung and Eric Harper and Eric Tramel and Erick Galinkin and Erik Pounds and Evan Briones and Evelina Bakhturina and Evgeny Tsykunov and Faisal Ladhak and Fay Wang and Fei Jia and Felipe Soares and Feng Chen and Ferenc Galko and Frank Sun and Frankie Siino and Gal Hubara Agam and Ganesh Ajjanagadde and Gantavya Bhatt and Gargi Prasad and George Armstrong and Gerald Shen and Gorkem Batmaz and Grigor Nalbandyan and Haifeng Qian and Harsh Sharma and Hayley Ross and Helen Ngo and Herbert Hum and Herman Sahota and Hexin Wang and Himanshu Soni and Hiren Upadhyay and Huizi Mao and Huy C Nguyen and Huy Q Nguyen and Iain Cunningham and Ido Galil and Ido Shahaf and Igor Gitman and Ilya Loshchilov and Itamar Schen and Itay Levy and Ivan Moshkov and Izik Golan and Izzy Putterman and Jan Kautz and Jane Polak Scowcroft and Jared Casper and Jatin Mitra and Jeffrey Glick and Jenny Chen and Jesse Oliver and Jian Zhang and Jiaqi Zeng and Jie Lou and Jimmy Zhang and Jinhang Choi and Jining Huang and Joey Conway and Joey Guman and John Kamalu and Johnny Greco and Jonathan Cohen and Joseph Jennings and Joyjit Daw and Julien Veron Vialard and Junkeun Yi and Jupinder Parmar and Kai Xu and Kan Zhu and Kari Briski and Katherine Cheung and Katherine Luna and Keith Wyss and Keshav Santhanam and Kevin Shih and Kezhi Kong and Khushi Bhardwaj and Kirthi Shankar and Krishna C. Puvvada and Krzysztof Pawelec and Kumar Anik and Lawrence McAfee and Laya Sleiman and Leon Derczynski and Li Ding and Lizzie Wei and Lucas Liebenwein and Luis Vega and Maanu Grover and Maarten Van Segbroeck and Maer Rodrigues de Melo and Mahdi Nazemi and Makesh Narsimhan Sreedhar and Manoj Kilaru and Maor Ashkenazi and Marc Romeijn and Marcin Chochowski and Mark Cai and Markus Kliegl and Maryam Moosaei and Matt Kulka and Matvei Novikov and Mehrzad Samadi and Melissa Corpuz and Mengru Wang and Meredith Price and Michael Andersch and Michael Boone and Michael Evans and Miguel Martinez and Mikail Khona and Mike Chrzanowski and Minseok Lee and Mohammad Dabbah and Mohammad Shoeybi and Mostofa Patwary and Nabin Mulepati and Najeeb Nabwani and Natalie Hereth and Nave Assaf and Negar Habibi and Neta Zmora and Netanel Haber and Nicola Sessions and Nidhi Bhatia and Nikhil Jukar and Nikki Pope and Nikolai Ludwig and Nima Tajbakhsh and Nir Ailon and Nirmal Juluru and Nishant Sharma and Oleksii Hrinchuk and Oleksii Kuchaiev and Olivier Delalleau and Oluwatobi Olabiyi and Omer Ullman Argov and Omri Puny and Oren Tropp and Ouye Xie and Parth Chadha and Pasha Shamis and Paul Gibbons and Pavlo Molchanov and Pawel Morkisz and Peter Dykas and Peter Jin and Pinky Xu and Piotr Januszewski and Pranav Prashant Thombre and Prasoon Varshney and Pritam Gundecha and Przemek Tredak and Qing Miao and Qiyu Wan and Rabeeh Karimi Mahabadi and Rachit Garg and Ran El-Yaniv and Ran Zilberstein and Rasoul Shafipour and Rich Harang and Rick Izzo and Rima Shahbazyan and Rishabh Garg and Ritika Borkar and Ritu Gala and Riyad Islam and Robert Hesse and Roger Waleffe and Rohit Watve and Roi Koren and Ruoxi Zhang and Russell Hewett and Russell J. Hewett and Ryan Prenger and Ryan Timbrook and Sadegh Mahdavi and Sahil Modi and Samuel Kriman and Sangkug Lim and Sanjay Kariyappa and Sanjeev Satheesh and Saori Kaji and Satish Pasumarthi and Saurav Muralidharan and Sean Narentharen and Sean Narenthiran and Seonmyeong Bak and Sergey Kashirsky and Seth Poulos and Shahar Mor and Shanmugam Ramasamy and Shantanu Acharya and Shaona Ghosh and Sharath Turuvekere Sreenivas and Shelby Thomas and Shiqing Fan and Shreya Gopal and Shrimai Prabhumoye and Shubham Pachori and Shubham Toshniwal and Shuoyang Ding and Siddharth Singh and Simeng Sun and Smita Ithape and Somshubra Majumdar and Soumye Singhal and Stas Sergienko and Stefania Alborghetti and Stephen Ge and Sugam Dipak Devare and Sumeet Kumar Barua and Suseella Panguluri and Suyog Gupta and Sweta Priyadarshi and Syeda Nahida Akter and Tan Bui and Teodor-Dumitru Ene and Terry Kong and Thanh Do and Tijmen Blankevoort and Tim Moon and Tom Balough and Tomer Asida and Tomer Bar Natan and Tomer Ronen and Tugrul Konuk and Twinkle Vashishth and Udi Karpas and Ushnish De and Vahid Noorozi and Vahid Noroozi and Venkat Srinivasan and Venmugil Elango and Victor Cui and Vijay Korthikanti and Vinay Rao and Vitaly Kurin and Vitaly Lavrukhin and Vladimir Anisimov and Wanli Jiang and Wasi Uddin Ahmad and Wei Du and Wei Ping and Wenfei Zhou and Will Jennings and William Zhang and Wojciech Prazuch and Xiaowei Ren and Yashaswi Karnati and Yejin Choi and Yev Meyer and Yi-Fu Wu and Yian Zhang and Yigong Qin and Ying Lin and Yonatan Geifman and Yonggan Fu and Yoshi Subara and Yoshi Suhara and Yubo Gao and Zach Moshe and Zhen Dong and Zhongbo Zhu and Zihan Liu and Zijia Chen and Zijie Yan},
      year={2025},
      eprint={2512.20856},
      archivePrefix={arXiv},
      primaryClass={cs.CL},
      url={https://arxiv.org/abs/2512.20856}, 
}

@misc{qwen3technicalreport,
      title={Qwen3 Technical Report}, 
      author={{Qwen Team}},
      year={2025},
      eprint={2505.09388},
      archivePrefix={arXiv},
      primaryClass={cs.CL},
      url={https://arxiv.org/abs/2505.09388}, 
}

@misc{yang2025gateddeltanetworksimproving,
      title={Gated Delta Networks: Improving Mamba2 with Delta Rule}, 
      author={Songlin Yang and Jan Kautz and Ali Hatamizadeh},
      year={2025},
      eprint={2412.06464},
      archivePrefix={arXiv},
      primaryClass={cs.CL},
      url={https://arxiv.org/abs/2412.06464}, 
}

@misc{li2025transmambaflexiblyswitchingtransformer,
      title={TransMamba: Flexibly Switching between Transformer and Mamba}, 
      author={Yixing Li and Ruobing Xie and Zhen Yang and Xingwu Sun and Shuaipeng Li and Weidong Han and Zhanhui Kang and Yu Cheng and Chengzhong Xu and Di Wang and Jie Jiang},
      year={2025},
      eprint={2503.24067},
      archivePrefix={arXiv},
      primaryClass={cs.LG},
      url={https://arxiv.org/abs/2503.24067}, 
}

@misc{openai2025gptoss120bgptoss20bmodel,
      title={{gpt-oss-120b \& gpt-oss-20b Model Card}}, 
      author={OpenAI and : and Sandhini Agarwal and Lama Ahmad and Jason Ai and Sam Altman and Andy Applebaum and Edwin Arbus and Rahul K. Arora and Yu Bai and Bowen Baker and Haiming Bao and Boaz Barak and Ally Bennett and Tyler Bertao and Nivedita Brett and Eugene Brevdo and Greg Brockman and Sebastien Bubeck and Che Chang and Kai Chen and Mark Chen and Enoch Cheung and Aidan Clark and Dan Cook and Marat Dukhan and Casey Dvorak and Kevin Fives and Vlad Fomenko and Timur Garipov and Kristian Georgiev and Mia Glaese and Tarun Gogineni and Adam Goucher and Lukas Gross and Katia Gil Guzman and John Hallman and Jackie Hehir and Johannes Heidecke and Alec Helyar and Haitang Hu and Romain Huet and Jacob Huh and Saachi Jain and Zach Johnson and Chris Koch and Irina Kofman and Dominik Kundel and Jason Kwon and Volodymyr Kyrylov and Elaine Ya Le and Guillaume Leclerc and James Park Lennon and Scott Lessans and Mario Lezcano-Casado and Yuanzhi Li and Zhuohan Li and Ji Lin and Jordan Liss and Lily and Liu and Jiancheng Liu and Kevin Lu and Chris Lu and Zoran Martinovic and Lindsay McCallum and Josh McGrath and Scott McKinney and Aidan McLaughlin and Song Mei and Steve Mostovoy and Tong Mu and Gideon Myles and Alexander Neitz and Alex Nichol and Jakub Pachocki and Alex Paino and Dana Palmie and Ashley Pantuliano and Giambattista Parascandolo and Jongsoo Park and Leher Pathak and Carolina Paz and Ludovic Peran and Dmitry Pimenov and Michelle Pokrass and Elizabeth Proehl and Huida Qiu and Gaby Raila and Filippo Raso and Hongyu Ren and Kimmy Richardson and David Robinson and Bob Rotsted and Hadi Salman and Suvansh Sanjeev and Max Schwarzer and D. Sculley and Harshit Sikchi and Kendal Simon and Karan Singhal and Yang Song and Dane Stuckey and Zhiqing Sun and Philippe Tillet and Sam Toizer and Foivos Tsimpourlas and Nikhil Vyas and Eric Wallace and Xin Wang and Miles Wang and Olivia Watkins and Kevin Weil and Amy Wendling and Kevin Whinnery and Cedric Whitney and Hannah Wong and Lin Yang and Yu Yang and Michihiro Yasunaga and Kristen Ying and Wojciech Zaremba and Wenting Zhan and Cyril Zhang and Brian Zhang and Eddie Zhang and Shengjia Zhao},
      year={2025},
      eprint={2508.10925},
      archivePrefix={arXiv},
      primaryClass={cs.CL},
      url={https://arxiv.org/abs/2508.10925}, 
}

@misc{gemmateam2025gemma3technicalreport,
      title={Gemma 3 Technical Report}, 
      author={{Gemma Team} and Aishwarya Kamath and Johan Ferret and Shreya Pathak and Nino Vieillard and Ramona Merhej and Sarah Perrin and Tatiana Matejovicova and Alexandre Ramé and Morgane Rivière and Louis Rouillard and Thomas Mesnard and Geoffrey Cideron and Jean-bastien Grill and Sabela Ramos and Edouard Yvinec and Michelle Casbon and Etienne Pot and Ivo Penchev and Gaël Liu and Francesco Visin and Kathleen Kenealy and Lucas Beyer and Xiaohai Zhai and Anton Tsitsulin and Robert Busa-Fekete and Alex Feng and Noveen Sachdeva and Benjamin Coleman and Yi Gao and Basil Mustafa and Iain Barr and Emilio Parisotto and David Tian and Matan Eyal and Colin Cherry and Jan-Thorsten Peter and Danila Sinopalnikov and Surya Bhupatiraju and Rishabh Agarwal and Mehran Kazemi and Dan Malkin and Ravin Kumar and David Vilar and Idan Brusilovsky and Jiaming Luo and Andreas Steiner and Abe Friesen and Abhanshu Sharma and Abheesht Sharma and Adi Mayrav Gilady and Adrian Goedeckemeyer and Alaa Saade and Alex Feng and Alexander Kolesnikov and Alexei Bendebury and Alvin Abdagic and Amit Vadi and András György and André Susano Pinto and Anil Das and Ankur Bapna and Antoine Miech and Antoine Yang and Antonia Paterson and Ashish Shenoy and Ayan Chakrabarti and Bilal Piot and Bo Wu and Bobak Shahriari and Bryce Petrini and Charlie Chen and Charline Le Lan and Christopher A. Choquette-Choo and CJ Carey and Cormac Brick and Daniel Deutsch and Danielle Eisenbud and Dee Cattle and Derek Cheng and Dimitris Paparas and Divyashree Shivakumar Sreepathihalli and Doug Reid and Dustin Tran and Dustin Zelle and Eric Noland and Erwin Huizenga and Eugene Kharitonov and Frederick Liu and Gagik Amirkhanyan and Glenn Cameron and Hadi Hashemi and Hanna Klimczak-Plucińska and Harman Singh and Harsh Mehta and Harshal Tushar Lehri and Hussein Hazimeh and Ian Ballantyne and Idan Szpektor and Ivan Nardini and Jean Pouget-Abadie and Jetha Chan and Joe Stanton and John Wieting and Jonathan Lai and Jordi Orbay and Joseph Fernandez and Josh Newlan and Ju-yeong Ji and Jyotinder Singh and Kat Black and Kathy Yu and Kevin Hui and Kiran Vodrahalli and Klaus Greff and Linhai Qiu and Marcella Valentine and Marina Coelho and Marvin Ritter and Matt Hoffman and Matthew Watson and Mayank Chaturvedi and Michael Moynihan and Min Ma and Nabila Babar and Natasha Noy and Nathan Byrd and Nick Roy and Nikola Momchev and Nilay Chauhan and Noveen Sachdeva and Oskar Bunyan and Pankil Botarda and Paul Caron and Paul Kishan Rubenstein and Phil Culliton and Philipp Schmid and Pier Giuseppe Sessa and Pingmei Xu and Piotr Stanczyk and Pouya Tafti and Rakesh Shivanna and Renjie Wu and Renke Pan and Reza Rokni and Rob Willoughby and Rohith Vallu and Ryan Mullins and Sammy Jerome and Sara Smoot and Sertan Girgin and Shariq Iqbal and Shashir Reddy and Shruti Sheth and Siim Põder and Sijal Bhatnagar and Sindhu Raghuram Panyam and Sivan Eiger and Susan Zhang and Tianqi Liu and Trevor Yacovone and Tyler Liechty and Uday Kalra and Utku Evci and Vedant Misra and Vincent Roseberry and Vlad Feinberg and Vlad Kolesnikov and Woohyun Han and Woosuk Kwon and Xi Chen and Yinlam Chow and Yuvein Zhu and Zichuan Wei and Zoltan Egyed and Victor Cotruta and Minh Giang and Phoebe Kirk and Anand Rao and Kat Black and Nabila Babar and Jessica Lo and Erica Moreira and Luiz Gustavo Martins and Omar Sanseviero and Lucas Gonzalez and Zach Gleicher and Tris Warkentin and Vahab Mirrokni and Evan Senter and Eli Collins and Joelle Barral and Zoubin Ghahramani and Raia Hadsell and Yossi Matias and D. Sculley and Slav Petrov and Noah Fiedel and Noam Shazeer and Oriol Vinyals and Jeff Dean and Demis Hassabis and Koray Kavukcuoglu and Clement Farabet and Elena Buchatskaya and Jean-Baptiste Alayrac and Rohan Anil and Dmitry and Lepikhin and Sebastian Borgeaud and Olivier Bachem and Armand Joulin and Alek Andreev and Cassidy Hardin and Robert Dadashi and Léonard Hussenot},
      year={2025},
      eprint={2503.19786},
      archivePrefix={arXiv},
      primaryClass={cs.CL},
      url={https://arxiv.org/abs/2503.19786}, 
}

@misc{irie2025blendingcomplementarymemorysystems,
      title={Blending Complementary Memory Systems in Hybrid Quadratic-Linear Transformers}, 
      author={Kazuki Irie and Morris Yau and Samuel J. Gershman},
      year={2025},
      eprint={2506.00744},
      archivePrefix={arXiv},
      primaryClass={cs.LG},
      url={https://arxiv.org/abs/2506.00744}, 
}

@misc{fang2025artificialhippocampusnetworksefficient,
      title={Artificial Hippocampus Networks for Efficient Long-Context Modeling}, 
      author={Yunhao Fang and Weihao Yu and Shu Zhong and Qinghao Ye and Xuehan Xiong and Lai Wei},
      year={2025},
      eprint={2510.07318},
      archivePrefix={arXiv},
      primaryClass={cs.CL},
      url={https://arxiv.org/abs/2510.07318}, 
}

@misc{vaswani2023attentionneed,
      title={Attention Is All You Need}, 
      author={Ashish Vaswani and Noam Shazeer and Niki Parmar and Jakob Uszkoreit and Llion Jones and Aidan N. Gomez and Lukasz Kaiser and Illia Polosukhin},
      year={2017},
      eprint={1706.03762},
      archivePrefix={arXiv},
      primaryClass={cs.CL},
      url={https://arxiv.org/abs/1706.03762}, 
}

@misc{hendrycks2023gaussianerrorlinearunits,
      title={Gaussian Error Linear Units (GELUs)}, 
      author={Dan Hendrycks and Kevin Gimpel},
      year={2016},
      eprint={1606.08415},
      archivePrefix={arXiv},
      primaryClass={cs.LG},
      url={https://arxiv.org/abs/1606.08415}, 
}

@misc{touvron2023llamaopenefficientfoundation,
      title={LLaMA: Open and Efficient Foundation Language Models}, 
      author={Hugo Touvron and Thibaut Lavril and Gautier Izacard and Xavier Martinet and Marie-Anne Lachaux and Timothée Lacroix and Baptiste Rozière and Naman Goyal and Eric Hambro and Faisal Azhar and Aurelien Rodriguez and Armand Joulin and Edouard Grave and Guillaume Lample},
      year={2023},
      eprint={2302.13971},
      archivePrefix={arXiv},
      primaryClass={cs.CL},
      url={https://arxiv.org/abs/2302.13971}, 
}

@misc{brown2020languagemodelsfewshotlearners,
      title={Language Models are Few-Shot Learners}, 
      author={Tom B. Brown and Benjamin Mann and Nick Ryder and Melanie Subbiah and Jared Kaplan and Prafulla Dhariwal and Arvind Neelakantan and Pranav Shyam and Girish Sastry and Amanda Askell and Sandhini Agarwal and Ariel Herbert-Voss and Gretchen Krueger and Tom Henighan and Rewon Child and Aditya Ramesh and Daniel M. Ziegler and Jeffrey Wu and Clemens Winter and Christopher Hesse and Mark Chen and Eric Sigler and Mateusz Litwin and Scott Gray and Benjamin Chess and Jack Clark and Christopher Berner and Sam McCandlish and Alec Radford and Ilya Sutskever and Dario Amodei},
      year={2020},
      eprint={2005.14165},
      archivePrefix={arXiv},
      primaryClass={cs.CL},
      url={https://arxiv.org/abs/2005.14165}, 
}

@misc{lozhkov2024fineweb-edu,
    author       = { Lozhkov, Anton and Ben Allal, Loubna and von Werra, Leandro and Wolf, Thomas },  
    title        = { FineWeb-Edu: the Finest Collection of Educational Content }, 
    year         = 2024,  
    url          = { https://huggingface.co/datasets/HuggingFaceFW/fineweb-edu },  
    doi          = { 10.57967/hf/2497 },
    publisher    = { Hugging Face }
}

@misc{Radford2019LanguageMA,
  title={Language Models are Unsupervised Multitask Learners},
  author={Alec Radford and Jeff Wu and Rewon Child and David Luan and Dario Amodei and Ilya Sutskever},
  year={2019},
  url={https://api.semanticscholar.org/CorpusID:160025533}
}

@misc{loshchilov2019decoupledweightdecayregularization,
      title={Decoupled Weight Decay Regularization}, 
      author={Ilya Loshchilov and Frank Hutter},
      year={2019},
      eprint={1711.05101},
      archivePrefix={arXiv},
      primaryClass={cs.LG},
      url={https://arxiv.org/abs/1711.05101}, 
}

@misc{paperno2016lambadadatasetwordprediction,
      title={The LAMBADA dataset: Word prediction requiring a broad discourse context}, 
      author={Denis Paperno and Germán Kruszewski and Angeliki Lazaridou and Quan Ngoc Pham and Raffaella Bernardi and Sandro Pezzelle and Marco Baroni and Gemma Boleda and Raquel Fernández},
      year={2016},
      eprint={1606.06031},
      archivePrefix={arXiv},
      primaryClass={cs.CL},
      url={https://arxiv.org/abs/1606.06031}, 
}

@misc{zellers2019hellaswagmachinereallyfinish,
      title={HellaSwag: Can a Machine Really Finish Your Sentence?}, 
      author={Rowan Zellers and Ari Holtzman and Yonatan Bisk and Ali Farhadi and Yejin Choi},
      year={2019},
      eprint={1905.07830},
      archivePrefix={arXiv},
      primaryClass={cs.CL},
      url={https://arxiv.org/abs/1905.07830}, 
}

@misc{bisk2019piqareasoningphysicalcommonsense,
      title={PIQA: Reasoning about Physical Commonsense in Natural Language}, 
      author={Yonatan Bisk and Rowan Zellers and Ronan Le Bras and Jianfeng Gao and Yejin Choi},
      year={2019},
      eprint={1911.11641},
      archivePrefix={arXiv},
      primaryClass={cs.CL},
      url={https://arxiv.org/abs/1911.11641}, 
}

@misc{clark2018thinksolvedquestionanswering,
      title={Think you have Solved Question Answering? Try ARC, the AI2 Reasoning Challenge}, 
      author={Peter Clark and Isaac Cowhey and Oren Etzioni and Tushar Khot and Ashish Sabharwal and Carissa Schoenick and Oyvind Tafjord},
      year={2018},
      eprint={1803.05457},
      archivePrefix={arXiv},
      primaryClass={cs.AI},
      url={https://arxiv.org/abs/1803.05457}, 
}

@misc{sakaguchi2019winograndeadversarialwinogradschema,
      title={WinoGrande: An Adversarial Winograd Schema Challenge at Scale}, 
      author={Keisuke Sakaguchi and Ronan Le Bras and Chandra Bhagavatula and Yejin Choi},
      year={2019},
      eprint={1907.10641},
      archivePrefix={arXiv},
      primaryClass={cs.CL},
      url={https://arxiv.org/abs/1907.10641}, 
}

@misc{mihaylov2018suitarmorconductelectricity,
      title={Can a Suit of Armor Conduct Electricity? A New Dataset for Open Book Question Answering}, 
      author={Todor Mihaylov and Peter Clark and Tushar Khot and Ashish Sabharwal},
      year={2018},
      eprint={1809.02789},
      archivePrefix={arXiv},
      primaryClass={cs.CL},
      url={https://arxiv.org/abs/1809.02789}, 
}

@misc{hoffmann2022trainingcomputeoptimallargelanguage,
      title={Training Compute-Optimal Large Language Models}, 
      author={Jordan Hoffmann and Sebastian Borgeaud and Arthur Mensch and Elena Buchatskaya and Trevor Cai and Eliza Rutherford and Diego de Las Casas and Lisa Anne Hendricks and Johannes Welbl and Aidan Clark and Tom Hennigan and Eric Noland and Katie Millican and George van den Driessche and Bogdan Damoc and Aurelia Guy and Simon Osindero and Karen Simonyan and Erich Elsen and Jack W. Rae and Oriol Vinyals and Laurent Sifre},
      year={2022},
      eprint={2203.15556},
      archivePrefix={arXiv},
      primaryClass={cs.CL},
      url={https://arxiv.org/abs/2203.15556}, 
}

@misc{choromanski2022rethinkingattentionperformers,
      title={Rethinking Attention with Performers}, 
      author={Krzysztof Choromanski and Valerii Likhosherstov and David Dohan and Xingyou Song and Andreea Gane and Tamas Sarlos and Peter Hawkins and Jared Davis and Afroz Mohiuddin and Lukasz Kaiser and David Belanger and Lucy Colwell and Adrian Weller},
      year={2022},
      eprint={2009.14794},
      archivePrefix={arXiv},
      primaryClass={cs.LG},
      url={https://arxiv.org/abs/2009.14794}, 
}

@misc{peng2021randomfeatureattention,
      title={Random Feature Attention}, 
      author={Hao Peng and Nikolaos Pappas and Dani Yogatama and Roy Schwartz and Noah A. Smith and Lingpeng Kong},
      year={2021},
      eprint={2103.02143},
      archivePrefix={arXiv},
      primaryClass={cs.CL},
      url={https://arxiv.org/abs/2103.02143}, 
}

@misc{xiong2021nystromformernystrombasedalgorithmapproximating,
      title={Nystr\"omformer: A Nystr\"om-Based Algorithm for Approximating Self-Attention}, 
      author={Yunyang Xiong and Zhanpeng Zeng and Rudrasis Chakraborty and Mingxing Tan and Glenn Fung and Yin Li and Vikas Singh},
      year={2021},
      eprint={2102.03902},
      archivePrefix={arXiv},
      primaryClass={cs.CL},
      url={https://arxiv.org/abs/2102.03902}, 
}

@misc{schlag2021lineartransformerssecretlyfast,
      title={Linear Transformers Are Secretly Fast Weight Programmers}, 
      author={Imanol Schlag and Kazuki Irie and Jürgen Schmidhuber},
      year={2021},
      eprint={2102.11174},
      archivePrefix={arXiv},
      primaryClass={cs.LG},
      url={https://arxiv.org/abs/2102.11174}, 
}

@misc{yang2025parallelizinglineartransformersdelta,
      title={Parallelizing Linear Transformers with the Delta Rule over Sequence Length}, 
      author={Songlin Yang and Bailin Wang and Yu Zhang and Yikang Shen and Yoon Kim},
      year={2025},
      eprint={2406.06484},
      archivePrefix={arXiv},
      primaryClass={cs.LG},
      url={https://arxiv.org/abs/2406.06484}, 
}

@misc{gu2022efficientlymodelinglongsequences,
      title={Efficiently Modeling Long Sequences with Structured State Spaces}, 
      author={Albert Gu and Karan Goel and Christopher Ré},
      year={2022},
      eprint={2111.00396},
      archivePrefix={arXiv},
      primaryClass={cs.LG},
      url={https://arxiv.org/abs/2111.00396}, 
}

@misc{hu2025combaimprovingbilinearrnns,
      title={Comba: Improving Bilinear RNNs with Closed-loop Control}, 
      author={Jiaxi Hu and Yongqi Pan and Jusen Du and Disen Lan and Xiaqiang Tang and Qingsong Wen and Yuxuan Liang and Weigao Sun},
      year={2025},
      eprint={2506.02475},
      archivePrefix={arXiv},
      primaryClass={cs.LG},
      url={https://arxiv.org/abs/2506.02475}, 
}

@misc{smith2023simplifiedstatespacelayers,
      title={Simplified State Space Layers for Sequence Modeling}, 
      author={Jimmy T. H. Smith and Andrew Warrington and Scott W. Linderman},
      year={2023},
      eprint={2208.04933},
      archivePrefix={arXiv},
      primaryClass={cs.LG},
      url={https://arxiv.org/abs/2208.04933}, 
}

@misc{zhang2025testtimetrainingright,
      title={Test-Time Training Done Right}, 
      author={Tianyuan Zhang and Sai Bi and Yicong Hong and Kai Zhang and Fujun Luan and Songlin Yang and Kalyan Sunkavalli and William T. Freeman and Hao Tan},
      year={2025},
      eprint={2505.23884},
      archivePrefix={arXiv},
      primaryClass={cs.LG},
      url={https://arxiv.org/abs/2505.23884}, 
}

@misc{tandon2025endtoendtesttimetraininglong,
      title={End-to-End Test-Time Training for Long Context}, 
      author={Arnuv Tandon and Karan Dalal and Xinhao Li and Daniel Koceja and Marcel Rød and Sam Buchanan and Xiaolong Wang and Jure Leskovec and Sanmi Koyejo and Tatsunori Hashimoto and Carlos Guestrin and Jed McCaleb and Yejin Choi and Yu Sun},
      year={2025},
      eprint={2512.23675},
      archivePrefix={arXiv},
      primaryClass={cs.LG},
      url={https://arxiv.org/abs/2512.23675}, 
}

@misc{behrouz2024titanslearningmemorizetest,
      title={Titans: Learning to Memorize at Test Time}, 
      author={Ali Behrouz and Peilin Zhong and Vahab Mirrokni},
      year={2024},
      eprint={2501.00663},
      archivePrefix={arXiv},
      primaryClass={cs.LG},
      url={https://arxiv.org/abs/2501.00663}, 
}

@misc{ibm_granite_2025,
  author= {{IBM Research}},
  title= {Granite 4.0 Models Documentation},
  howpublished = {\url{https://www.ibm.com/granite/docs/models/granite}},
  year= {2025},
  note= {Accessed: 2025-02-06}
}

@misc{beltagy2020longformerlongdocumenttransformer,
      title={Longformer: The Long-Document Transformer}, 
      author={Iz Beltagy and Matthew E. Peters and Arman Cohan},
      year={2020},
      eprint={2004.05150},
      archivePrefix={arXiv},
      primaryClass={cs.CL},
      url={https://arxiv.org/abs/2004.05150}, 
}

@misc{yuan2025nativesparseattentionhardwarealigned,
      title={Native Sparse Attention: Hardware-Aligned and Natively Trainable Sparse Attention}, 
      author={Jingyang Yuan and Huazuo Gao and Damai Dai and Junyu Luo and Liang Zhao and Zhengyan Zhang and Zhenda Xie and Y. X. Wei and Lean Wang and Zhiping Xiao and Yuqing Wang and Chong Ruan and Ming Zhang and Wenfeng Liang and Wangding Zeng},
      year={2025},
      eprint={2502.11089},
      archivePrefix={arXiv},
      primaryClass={cs.CL},
      url={https://arxiv.org/abs/2502.11089}, 
}

@misc{fedus2022switchtransformersscalingtrillion,
      title={Switch Transformers: Scaling to Trillion Parameter Models with Simple and Efficient Sparsity}, 
      author={William Fedus and Barret Zoph and Noam Shazeer},
      year={2022},
      eprint={2101.03961},
      archivePrefix={arXiv},
      primaryClass={cs.LG},
      url={https://arxiv.org/abs/2101.03961}, 
}

@misc{shazeer2017outrageouslylargeneuralnetworks,
      title={Outrageously Large Neural Networks: The Sparsely-Gated Mixture-of-Experts Layer}, 
      author={Noam Shazeer and Azalia Mirhoseini and Krzysztof Maziarz and Andy Davis and Quoc Le and Geoffrey Hinton and Jeff Dean},
      year={2017},
      eprint={1701.06538},
      archivePrefix={arXiv},
      primaryClass={cs.LG},
      url={https://arxiv.org/abs/1701.06538}, 
}

@misc{wang2024auxiliarylossfreeloadbalancingstrategy,
      title={Auxiliary-Loss-Free Load Balancing Strategy for Mixture-of-Experts}, 
      author={Lean Wang and Huazuo Gao and Chenggang Zhao and Xu Sun and Damai Dai},
      year={2024},
      eprint={2408.15664},
      archivePrefix={arXiv},
      primaryClass={cs.LG},
      url={https://arxiv.org/abs/2408.15664}, 
}

@misc{lin2024momaefficientearlyfusionpretraining,
      title={MoMa: Efficient Early-Fusion Pre-training with Mixture of Modality-Aware Experts}, 
      author={Xi Victoria Lin and Akshat Shrivastava and Liang Luo and Srinivasan Iyer and Mike Lewis and Gargi Ghosh and Luke Zettlemoyer and Armen Aghajanyan},
      year={2024},
      eprint={2407.21770},
      archivePrefix={arXiv},
      primaryClass={cs.AI},
      url={https://arxiv.org/abs/2407.21770}, 
}

@misc{wu2024multiheadmixtureofexperts,
      title={Multi-Head Mixture-of-Experts}, 
      author={Xun Wu and Shaohan Huang and Wenhui Wang and Furu Wei},
      year={2024},
      eprint={2404.15045},
      archivePrefix={arXiv},
      primaryClass={cs.CL},
      url={https://arxiv.org/abs/2404.15045}, 
}

@misc{zhang2022mixtureattentionheadsselecting,
      title={Mixture of Attention Heads: Selecting Attention Heads Per Token}, 
      author={Xiaofeng Zhang and Yikang Shen and Zeyu Huang and Jie Zhou and Wenge Rong and Zhang Xiong},
      year={2022},
      eprint={2210.05144},
      archivePrefix={arXiv},
      primaryClass={cs.CL},
      url={https://arxiv.org/abs/2210.05144}, 
}

@misc{roy2025fastsimplex2simplicialattention,
      title={Fast and Simplex: 2-Simplicial Attention in Triton}, 
      author={Aurko Roy and Timothy Chou and Sai Surya Duvvuri and Sijia Chen and Jiecao Yu and Xiaodong Wang and Manzil Zaheer and Rohan Anil},
      year={2025},
      eprint={2507.02754},
      archivePrefix={arXiv},
      primaryClass={cs.LG},
      url={https://arxiv.org/abs/2507.02754}, 
}

@misc{wang2025testtimeregressionunifyingframework,
      title={Test-time regression: a unifying framework for designing sequence models with associative memory}, 
      author={Ke Alexander Wang and Jiaxin Shi and Emily B. Fox},
      year={2025},
      eprint={2501.12352},
      archivePrefix={arXiv},
      primaryClass={cs.LG},
      url={https://arxiv.org/abs/2501.12352}, 
}

@article{trockman2024mimetic,
  title={Mimetic initialization helps state space models learn to recall},
  author={Trockman, Asher and Harutyunyan, Hrayr and Kolter, J Zico and Kumar, Sanjiv and Bhojanapalli, Srinadh},
  journal={arXiv preprint arXiv:2410.11135},
  year={2024}
}

@misc{lahoti2026mamba3improvedsequencemodeling,
      title={Mamba-3: Improved Sequence Modeling using State Space Principles}, 
      author={Aakash Lahoti and Kevin Y. Li and Berlin Chen and Caitlin Wang and Aviv Bick and J. Zico Kolter and Tri Dao and Albert Gu},
      year={2026},
      eprint={2603.15569},
      archivePrefix={arXiv},
      primaryClass={cs.LG},
      url={https://arxiv.org/abs/2603.15569}, 
}

@misc{shi2024wonderfulmatricescombiningefficient,
      title={Wonderful Matrices: Combining for a More Efficient and Effective Foundation Model Architecture}, 
      author={Jingze Shi and Bingheng Wu},
      year={2024},
      eprint={2412.11834},
      archivePrefix={arXiv},
      primaryClass={cs.LG},
      url={https://arxiv.org/abs/2412.11834}, 
}

@misc{hwang2025dynamicchunkingendtoendhierarchical,
      title={Dynamic Chunking for End-to-End Hierarchical Sequence Modeling}, 
      author={Sukjun Hwang and Brandon Wang and Albert Gu},
      year={2025},
      eprint={2507.07955},
      archivePrefix={arXiv},
      primaryClass={cs.LG},
      url={https://arxiv.org/abs/2507.07955}, 
}

@misc{lufkin2026hybridassociativememories,
      title={Hybrid Associative Memories}, 
      author={Leon Lufkin and Tomás Figliolia and Beren Millidge and Kamesh Krishnamurthy},
      year={2026},
      eprint={2603.22325},
      archivePrefix={arXiv},
      primaryClass={cs.LG},
      url={https://arxiv.org/abs/2603.22325}, 
}

@misc{zhang2026sla2sparselinearattentionlearnable,
      title={SLA2: Sparse-Linear Attention with Learnable Routing and QAT}, 
      author={Jintao Zhang and Haoxu Wang and Kai Jiang and Kaiwen Zheng and Youhe Jiang and Ion Stoica and Jianfei Chen and Jun Zhu and Joseph E. Gonzalez},
      year={2026},
      eprint={2602.12675},
      archivePrefix={arXiv},
      primaryClass={cs.LG},
      url={https://arxiv.org/abs/2602.12675}, 
}

@misc{liu2024longhornstatespacemodels,
      title={Longhorn: State Space Models are Amortized Online Learners}, 
      author={Bo Liu and Rui Wang and Lemeng Wu and Yihao Feng and Peter Stone and Qiang Liu},
      year={2024},
      eprint={2407.14207},
      archivePrefix={arXiv},
      primaryClass={cs.LG},
      url={https://arxiv.org/abs/2407.14207}, 
}

@misc{hsieh2024rulerwhatsrealcontext,
      title={RULER: What's the Real Context Size of Your Long-Context Language Models?}, 
      author={Cheng-Ping Hsieh and Simeng Sun and Samuel Kriman and Shantanu Acharya and Dima Rekesh and Fei Jia and Yang Zhang and Boris Ginsburg},
      year={2024},
      eprint={2404.06654},
      archivePrefix={arXiv},
      primaryClass={cs.CL},
      url={https://arxiv.org/abs/2404.06654}, 
}

@misc{arora2024justreadtwiceclosing,
      title={Just read twice: closing the recall gap for recurrent language models}, 
      author={Simran Arora and Aman Timalsina and Aaryan Singhal and Benjamin Spector and Sabri Eyuboglu and Xinyi Zhao and Ashish Rao and Atri Rudra and Christopher Ré},
      year={2024},
      eprint={2407.05483},
      archivePrefix={arXiv},
      primaryClass={cs.CL},
      url={https://arxiv.org/abs/2407.05483}, 
}

@misc{arora2025languagemodelsenablesimple,
      title={Language Models Enable Simple Systems for Generating Structured Views of Heterogeneous Data Lakes}, 
      author={Simran Arora and Brandon Yang and Sabri Eyuboglu and Avanika Narayan and Andrew Hojel and Immanuel Trummer and Christopher Ré},
      year={2025},
      eprint={2304.09433},
      archivePrefix={arXiv},
      primaryClass={cs.CL},
      url={https://arxiv.org/abs/2304.09433}, 
}

@inproceedings{lockard-etal-2019-openceres,
    title = "{O}pen{C}eres: {W}hen Open Information Extraction Meets the Semi-Structured Web",
    author = "Lockard, Colin  and
      Shiralkar, Prashant  and
      Dong, Xin Luna",
    editor = "Burstein, Jill  and
      Doran, Christy  and
      Solorio, Thamar",
    booktitle = "Proceedings of the 2019 Conference of the North {A}merican Chapter of the Association for Computational Linguistics: Human Language Technologies, Volume 1 (Long and Short Papers)",
    month = jun,
    year = "2019",
    address = "Minneapolis, Minnesota",
    publisher = "Association for Computational Linguistics",
    url = "https://aclanthology.org/N19-1309/",
    doi = "10.18653/v1/N19-1309",
    pages = "3047--3056"
}

@misc{rajpurkar2018know,
    title={Know What You Don't Know: Unanswerable Questions for SQuAD},
    author={Pranav Rajpurkar and Robin Jia and Percy Liang},
    year={2018},
    eprint={1806.03822},
    archivePrefix={arXiv},
    primaryClass={cs.CL}
}

@misc{joshi2017triviaqalargescaledistantly,
      title={TriviaQA: A Large Scale Distantly Supervised Challenge Dataset for Reading Comprehension}, 
      author={Mandar Joshi and Eunsol Choi and Daniel S. Weld and Luke Zettlemoyer},
      year={2017},
      eprint={1705.03551},
      archivePrefix={arXiv},
      primaryClass={cs.CL},
      url={https://arxiv.org/abs/1705.03551}, 
}

@article{kwiatkowski-etal-2019-natural,
    title = "Natural Questions: A Benchmark for Question Answering Research",
    author = "Kwiatkowski, Tom  and
      Palomaki, Jennimaria  and
      Redfield, Olivia  and
      Collins, Michael  and
      Parikh, Ankur  and
      Alberti, Chris  and
      Epstein, Danielle  and
      Polosukhin, Illia  and
      Devlin, Jacob  and
      Lee, Kenton  and
      Toutanova, Kristina  and
      Jones, Llion  and
      Kelcey, Matthew  and
      Chang, Ming-Wei  and
      Dai, Andrew M.  and
      Uszkoreit, Jakob  and
      Le, Quoc  and
      Petrov, Slav",
    editor = "Lee, Lillian  and
      Johnson, Mark  and
      Roark, Brian  and
      Nenkova, Ani",
    journal = "Transactions of the Association for Computational Linguistics",
    volume = "7",
    year = "2019",
    address = "Cambridge, MA",
    publisher = "MIT Press",
    url = "https://aclanthology.org/Q19-1026/",
    doi = "10.1162/tacl_a_00276",
    pages = "452--466"
}

@misc{dua2019dropreadingcomprehensionbenchmark,
      title={DROP: A Reading Comprehension Benchmark Requiring Discrete Reasoning Over Paragraphs}, 
      author={Dheeru Dua and Yizhong Wang and Pradeep Dasigi and Gabriel Stanovsky and Sameer Singh and Matt Gardner},
      year={2019},
      eprint={1903.00161},
      archivePrefix={arXiv},
      primaryClass={cs.CL},
      url={https://arxiv.org/abs/1903.00161}, 
}

@misc{yang2024gatedlinearattentiontransformers,
      title={Gated Linear Attention Transformers with Hardware-Efficient Training}, 
      author={Songlin Yang and Bailin Wang and Yikang Shen and Rameswar Panda and Yoon Kim},
      year={2024},
      eprint={2312.06635},
      archivePrefix={arXiv},
      primaryClass={cs.LG},
      url={https://arxiv.org/abs/2312.06635}, 
}

@misc{dao2022flashattentionfastmemoryefficientexact,
      title={FlashAttention: Fast and Memory-Efficient Exact Attention with IO-Awareness}, 
      author={Tri Dao and Daniel Y. Fu and Stefano Ermon and Atri Rudra and Christopher Ré},
      year={2022},
      eprint={2205.14135},
      archivePrefix={arXiv},
      primaryClass={cs.LG},
      url={https://arxiv.org/abs/2205.14135}, 
}
\clearpage
\appendix
\section{Additional Related Work}
\label{sec:additional-related}
\paragraph{Linear Layers.}
There has been a growing body of work focused on mitigating the memory and compute requirements of the softmax attention mechanism.
These alternatives either aim to approximate the softmax attention mechanism~\citep{katharopoulos2020transformersrnnsfastautoregressive,choromanski2022rethinkingattentionperformers,peng2021randomfeatureattention,xiong2021nystromformernystrombasedalgorithmapproximating,hua2022transformerqualitylineartime,beltagy2020longformerlongdocumenttransformer} or are inspired by other mechanisms.
Many recent linear attention style models can be viewed as fast weight programmers~\citep{yang2025gateddeltanetworksimproving,schlag2021lineartransformerssecretlyfast,yang2025parallelizinglineartransformersdelta,hu2025combaimprovingbilinearrnns}, and the traditional state-space model (SSM) from control theory has also influenced many popular works~\citep{gu2024mambalineartimesequencemodeling,dao2024transformersssmsgeneralizedmodels,gu2022efficientlymodelinglongsequences,smith2023simplifiedstatespacelayers}.
\cite{trockman2024mimetic} showed that initializing such SSM layers to mimic attention layers improves performance on recall tasks, hinting at representational compatibility between the two layer types. 
Beyond linear attention and SSMs, test-time training (TTT) layers redefine the state as adjustable model weights during inference instead of an external tensor~\citep{zhang2025testtimetrainingright,tandon2025endtoendtesttimetraininglong,behrouz2024titanslearningmemorizetest}.
While these models do reduce the overall overhead needed for training and deployment, there exists a fundamental trade-off: the fixed-state size of linear RNNs forces lossy compression.
This weakness becomes apparent in retrieval or in-context learning heavy tasks~\citep{waleffe2024empiricalstudymambabasedlanguage,arora2025simplelinearattentionlanguage}.

\paragraph{Hybrid Models.} Given efficiency benefits and strong modeling capabilities, linear layers are increasingly incorporated alongside standard softmax attention within hybrid models to mitigate this weakness~\citep{nvidia2025nvidianemotron3efficient,qwen3technicalreport,openai2025gptoss120bgptoss20bmodel,kimiteam2025kimilinearexpressiveefficient,ibm_granite_2025,gemmateam2025gemma3technicalreport}.
Current hybrid model design can mainly be split into two approaches: inter-layer and intra-layer.
The majority of current hybrid models follow the inter-layer format where different mixer mechanisms are interleaved on a layer basis where each layer only utilizes a single sequence mixer, e.g., three Mamba layers followed by one softmax attention layer.
Intra-layer designs integrate different types of mixer within a single layer or block.
For instance, Hymba~\citep{dong2024hymbahybridheadarchitecturesmall} utilizes sliding-window softmax attention and Mamba-2 in parallel within the same layer while maintaining completely separate parameters for both mixers prior to merging their outputs.
\citet{irie2025blendingcomplementarymemorysystems, fang2025artificialhippocampusnetworksefficient} utilize linear models as a mechanism to retain information discarded from the context of sliding-window, and thus can also be viewed as a variant of intra-layer hybrid models.
Native Sparse Attention enables some sequence-level adaptivity by selecting the context processed by its attention pathways, but its per-token compute budget is largely fixed based on the configured sparsity pattern~\citep{yuan2025nativesparseattentionhardwarealigned}.
More broadly, most current hybrid models either use a static mixture of mixer types or restrict adaptivity to routing among attention pathways, dedicating a ``fixed'' architecture to each token, which can be computationally inefficient given the heterogeneous nature of sequences and queries.
In contrast, \antelope{} enables the sequence mixing mechanism to be determined at each timestep.

\paragraph{Routing Mechanisms.}
The potential and usage of the \antelope\ block has parallels with routing-based layers and mechanisms.
While the mixer selection was chosen manually in our paper, the ability to utilize both mechanisms freely enables the ability for a token to decide the selected mixer, similar to current mixture-of-expert routing~\citep{fedus2022switchtransformersscalingtrillion,shazeer2017outrageouslylargeneuralnetworks,wang2024auxiliarylossfreeloadbalancingstrategy}.
While general MoE layers typically utilize homogeneous experts, \citet{lin2024momaefficientearlyfusionpretraining} partitions experts based on modality, similar to how our approach partitions the sequence mixer based on quadratic versus linear mechanisms.
Beyond channel mixers, i.e., MLPs, routing has also enabled the selection of certain attention heads within the sequence mixer~\citep{wu2024multiheadmixtureofexperts,zhang2022mixtureattentionheadsselecting}.
With the popularization of linear models with fixed-size states, methods have been proposed that route tokens to specific states to prevent overwriting of past information in other states~\citep{du2025momlinearsequencemodeling}.

\section{\antelope\ Architecture Details and Additional Ablations}
\label{sec:additional-ablations}
\paragraph{Mamba-2 \antelope.}
The RMSNorm used in the Mamba-2 variant of the \antelope\ block uses vanilla RMSNorm applied after the element-wise gating, following \citet{dao2024transformersssmsgeneralizedmodels}.
We find that GroupNorm~\citep{wu2018groupnormalization}, used in Hymba~\citep{dong2024hymbahybridheadarchitecturesmall}, harms the shared block's Mamba-2 performance, but find that GroupNorm can be utilized if the Mamba-2 mixer were replaced with one that operates well with GroupNorm, e.g., Gated DeltaNet~\citep{yang2025gateddeltanetworksimproving}.
Unlike standard Mamba blocks, we also remove the activations after the convolution to maintain better consistency with the standard Transformer design, and find this does not significantly impact performance.
In addition, we find that applying the RoPE positional embedding globally to queries and to the shared key empirically helps both mixers' performance (\Cref{tab:rope_ablation}).
Notably, integrating RoPE for both mixers improves Mamba-2's perplexity more significantly than it does for self-attention.
While it is unclear why this is the case, this finding may be attributed to the beneficial nature of operating with complex SSMs~\citep{lahoti2026mamba3improvedsequencemodeling} or incorporating direct positional encoding information rather than the decay's indirect information~\citep{shi2024wonderfulmatricescombiningefficient}.
\begin{table}[h]
    \small
    \centering
    \caption{
        Applying RoPE to both self-attention and Mamba queries and keys improves performance compared to applying it solely to self-attention.
        This holds true regardless of whether a short convolution is used on the key-value components.
    }
    \begin{tabular}{c c S S}
        \toprule
        \textbf{Conv?} & \textbf{RoPE Applied} & {Attn ppl $\downarrow$} & {Mamba-2 ppl $\downarrow$} \\
        \midrule
        \multirow{2}{*}{\shortstack[l]{\cmark}} 
        & Both & 15.54 & 15.54 \\
        & Attn & 15.59 & 15.70 \\
        \midrule
        \multirow{2}{*}{\shortstack[l]{\xmark}} 
        & Both & 15.85 & 15.94 \\
        & Attn & 16.05 & 16.03 \\
        \bottomrule
    \end{tabular}
    \label{tab:rope_ablation}
\end{table}
\newline
\paragraph{GDN \antelope.}
Following~\citep{yang2025gateddeltanetworksimproving}, the pre-output projection norm uses the pre-gate, GroupNorm design.
GDN requires the L2 normalization of queries and keys, which we find empirically hurts the performance of the attention mode significantly when applied (\Cref{tab:gdn_norm_ablation}).
Thus, we only apply L2 normalization to the queries and keys passed into the GDN mixer.
Unlike the Mamba-2 variant of \antelope, the post-short convolution activation is important for modeling and retrieval performance, thus we add the standard SiLU activation to both mixers' queries and the shared keys and values (\Cref{tab:gdn_rope_act_ablation}).
Furthermore, rotary embeddings are only applied to the attention mixer's query and keys, as the application to GDN hurts retrieval performance.
The mixer-specific application of the architectural components, such as the embeddings and norms, can be subsumed by the mixer kernels with little loss in efficiency, as the main projection still shared, but we leave the implementation for future work.
\begin{table}[ht]
    \small
    \centering
    \caption{
    GDN requires L2 norm applied to its query and key, but we find not normalizing those of the attention mixer leads to the best results.
    }
    \begin{tabular}{l S S}
        \toprule
        \textbf{L2 Norm on Attn} & {Attn ppl $\downarrow$} & {GDN ppl $\downarrow$} \\
        \midrule
        Query + Key & 28.68 & 16.16 \\
        Key & 15.18 & 15.32 \\
        None & 15.08 & 15.06 \\
        \bottomrule
    \end{tabular}
    \label{tab:gdn_norm_ablation}
\end{table}

\begin{table}[ht]
    \small
    \centering
    \caption{
    \textbf{Left:} SiLU activation on the keys and values is critical for performance.
    While applying RoPE only in the attention mixer performs similar to that of applying RoPE in both attention and GDN for pretraining loss, it performs better on retrieval-based tasks.
    \textbf{Right:} Using attention-only RoPE, we find that applying the SiLU activation on the queries for both attention and GDN leads to strong performance and better retrieval than other configurations.
    }
    \resizebox{\textwidth}{!}{
        \begin{tabular}{l l S S}
            \toprule
            \textbf{RoPE Applied} & \textbf{Shared KV Act.} & {Attn ppl $\downarrow$} & {GDN ppl $\downarrow$} \\
            \midrule
            \multirow{2}{*}{Both} & SiLU & 15.08 & 15.06 \\
            & Id. & 15.21 & 15.30 \\
            \midrule
            \multirow{2}{*}{Attn} & SiLU & 15.09 & 15.07 \\
            & Id. & 15.23 & 15.29 \\
            \bottomrule
        \end{tabular}

        \hspace{0.25cm} 

        \begin{tabular}{l l S S}
            \toprule
            \textbf{Attn Q act.} & \textbf{GDN Q act.} & {Attn ppl $\downarrow$} & {GDN ppl $\downarrow$} \\
            \midrule
            \multirow{2}{*}{SiLU} & SiLU & 15.09 & 15.07 \\
            & Id. & 15.31 & 15.28 \\
            \midrule
            \multirow{2}{*}{Id.} & SiLU & 15.04 & 15.06 \\
            & Id. & 15.23 & 15.26 \\
            \bottomrule
        \end{tabular}
    }
    \label{tab:gdn_rope_act_ablation}
\end{table}

\subsection{Pseudocode}
\begin{figure}[!th]
\centering
\begin{minipage}{\linewidth}

Sample implementations can be found for GatedRMSNorm\footnote{\,\url{https://github.com/state-spaces/mamba/blob/main/mamba_ssm/ops/triton/layernorm_gated.py}}
and ShortConv\footnote{\,\url{https://github.com/Dao-AILab/causal-conv1d}}.
\begin{lstlisting}[
style=pseudocode,
caption={Abstracted pseudocode for the Mamba-2 variant of the \antelope{} shared block. We omit chunk-based mixed-mode forwarding, the joint state update, and implementation details for clarity and focus on the general representations and components.},
label={lst:oryx_clean_pseudocode}
]
# Learnable parameters/modules:
#   W_K, W_V, W_G, W_O, W_Q_attn, W_Q_mamba, W_sup_mamba
#   ShortConv, GatedRMSNorm 

def OryxBlock(X, mode):
    K = X @ W_K
    V = X @ W_V
    G = SiLU(X @ W_G)

    K, V = ShortConv(K, V)

    if mode == "attention":
        Q = RoPE(X @ W_Q_attn)
        O = SoftmaxAttention(
            Q=Q,
            K=RoPE(K),
            V=V,
        )

    elif mode == "mamba2":
        Q = RoPE(X @ W_Q_mamba)
        O = Mamba2Mixer(
            x=V,
            B=RoPE(K),
            C=Q,
            params=W_sup_mamba,
        )

    Y = GatedRMSNorm(input=O, gate=G) @ W_O
    return Y
\end{lstlisting}
\end{minipage}
\end{figure}

\section{General FLOPs Analysis}
\label{sec:compute-analysis}
In this section, we provide a general FLOPs analysis of a forward pass of the \antelope{} sequence mixer algorithm and compare it against softmax attention and the Mamba-2 recurrent mechanism.
The analysis is intended to compare asymptotic costs of the base sequence-mixing operation under a simplified chunked implementation, thus, we omit the compute associated with the weight projections and additional components, e.g., gate, and hardware-specific effects.
As a refresher, modern recurrent linear models utilize a chunk forward algorithm during training or prefill, in which the total sequence length $T$ is partitioned into chunks of size $C$ and the outputs and states are calculated in parallel.
Thus, $\bmQ$ is partitioned into $[\bmQ_1;\cdots;\bmQ_{T/C}]$ where $\bmQ_i \in \mathbb{R}^{C \times D_k}$, and the same is applied to $\bmK, \bmV$.
The comprehensive algorithm can be found at~\citet{yang2024gatedlinearattentiontransformers,dao2024transformersssmsgeneralizedmodels}.
\paragraph{Linear RNN (Mamba-2) FLOPs count.}
The forward pass of a chunked linear RNN can generally be decomposed into four steps. We mainly consider the large matrix multiplication operations and ignore smaller element-wise multiplication operations.
\begin{enumerate}
    \item \textbf{Chunk state:} Here, each chunk $i$ computes its contribution to the recurrent state $\bmS_i\in\mathbb{R}^{D_k\times D_v}$ via $\bmK_i^\top \bmV_i$ which results in $2CD_kD_v$ FLOPs per chunk and $2TD_kD_v$ for the entire sequence.
    \item \textbf{State passing:} The chunk states are updated to incorporate prior state information via a scan on the $T/C$ total states of size $D_k\times D_v$.
    This results in approximately $2TD_kD_v/C$ FLOPs.
    \item \textbf{Intra-chunk output:} Here, the output from the intra-chunk interactions are calculated via $(\bmL_i \circ \bmQ_i \bmK_i^\top) \bmV_i$ which results in $2C^2D_k + 2C^2D_v$ per chunk, when ignoring the mask. The total FLOPs for all chunks are then $2TCD_k + 2TCD_v$.
    \item \textbf{Inter-chunk output:} Finally, the output from the cumulative prior hidden states must be calculated and added to the output arising from the intra-chunk calculations.
    Here, the $\bmQ_i \bmS_{i-1}$ calculation results in $2CD_kD_v$ per chunk and $2TD_kD_v$ in total.
    The summation of the inter-chunk and intra-chunk outputs results in $TP$ FLOPs, which is negligible, thus ignored.
\end{enumerate}
In total, the overall compute required for a single forward pass is approximately $4TD_kD_v + 2TC(D_k+D_v) + 2TD_kD_v/C$ which can be simplified to $8TC^2 + 2TC \approx 8TC^2$ when assuming $C=D_k=D_v$.
\paragraph{Softmax Attention FLOPs count.}
The standard causal softmax attention mechanism computes the quadratic interactions among the sequence constrained by a causal mask.
While hardware-aware implementations exist~\citep{dao2022flashattentionfastmemoryefficientexact}, they do not decrease the overall FLOPs count.
Similar to above, we ignore operations such as softmax and mainly focus on matmuls for clarity.
For the score computation $\bmQ\bmK^\top$, as only $T(T+1)/2$ pairs are valid due to causality, the total FLOPs required is $T(T+1)D_k$.
The aggregation of $\bmV$ follows the same argument, resulting in a $T(T+1)D_v$ FLOPs.
Thus, the total FLOPs required is $T(T+1)(D_k+D_v)$ which is around $2T^2C$ when assuming $C=D_k=D_v$.
\paragraph{\antelope{} FLOPs count.}
As \antelope{} uses shared key-value representations to update both the attention KV cache and linear state, the linear state needs to be updated at all instances.
However, because the mixer selection determines the output, the FLOPs required for generating the output depends on $\delta$ which we will assign as the probability that a chunk is routed to the linear mechanism.
Thus, we can consider the three aspects of \antelope{}'s compute.
\begin{enumerate}
    \item \textbf{Constant linear state update:} All chunks of the sequence need to update the state, resulting in $2TD_kD_v + 2TD_kD_v/C$ total FLOPs when accounting for the chunk state and state passing portions of the linear model.
    \item \textbf{Linear output:} As only $\delta$ chunks are assigned to the linear mixer and thus require computation, the overall number of FLOPs used is approximately $\delta(2TCD_k + 2TCD_v + 2TD_kD_v)$ when accounting for the intra- and inter-chunk operations.
    \item \textbf{Attention output:} Here, we make the assumption that the routing $\delta$ selects chunks uniformly at random to help with analysis.
    Under this simplifying condition, the computation cost in expectation is $(1-\delta)T(T+1)(D_k+D_v)$.
\end{enumerate}
When combined, the total compute required for an \antelope{} forward pass is $2TD_kD_v + 2TD_kD_v/C +\delta(2TCD_k+2TCD_v+2TD_kD_v) + (1-\delta)T(T+1)(D_k+D_v)$ which is approximately $2TC^2(1+3\delta)+2(1-\delta)T^2C$ when assuming $C=D_k=D_v$ and ignoring lower order terms.
\paragraph{Compute Tradeoff.}
When comparing the general FLOPs usage of \antelope{} and pure attention, \antelope{} uses fewer FLOPs when $(1 - \delta)+(1+3\delta)C/T<1$.
Rearranging, we find that in general, if $T > (\frac{1}{\delta} + 3)C$, the shared block utilizes less compute than pure attention.
This result is intuitive as a large $\delta$ would reduce the majority of the quadratic attention compute required, while a small $\delta$ would require most of the attention computation plus the recurrent state overhead.
In our one-to-three attention-to-linear split setting with $C=128$, the \antelope{} block would utilize less FLOPs than that of softmax attention if the context length were 2048 ($T=2048\geq555$).

\section{Additional Mode Switching Results}
\label{sec:additional-mode-switching}
\begin{figure}[h]
    \centering
    \includegraphics[width=\textwidth]{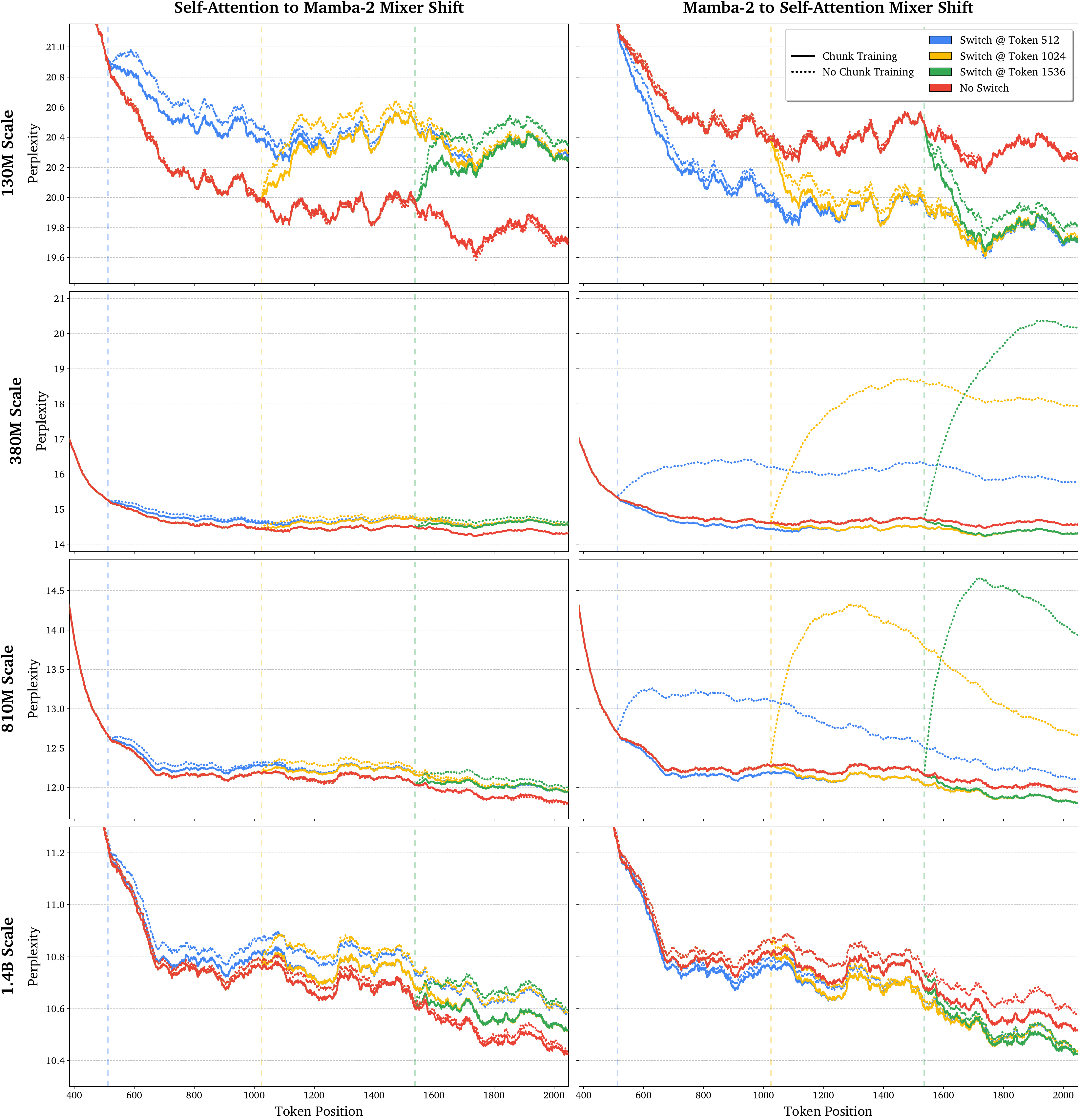}
    \caption{Smoothed perplexity across token index position for \antelope{}-TM models at all four scales trained with and without chunk-level switching.}
    \label{fig:scaling-mode-training}
\end{figure}

For instance, \antelope{}-TM models trained with sequence-level switching display the ability to freely switch from softmax-attention to Mamba-2 processing and vice versa without performance degradation at the 125M and 1.4B scale.
However, the 350M and 760M models can only switch from softmax-attention to Mamba-2 without issue: switching from Mamba-2 to softmax-attention leads to notable performance degradations as measured by a spike in token-index perplexity.
We display the perplexity of all Chinchilla-token trained models in \Cref{fig:scaling-mode-training}.

The cause of this divergence remains unclear, although our ablations suggest this is not an architectural issue.
Reducing learning rate mitigates the divergence and increasing training does not solve the issue.
We hypothesize this is due to the switch from softmax-attention to Mamba-2 involves switching from a more expressive to less expressive mixer which is easier than vice versa, which is why we see the ability appear naturally more often.

\begin{figure}[h]
    \centering
    \includegraphics[width=\textwidth]{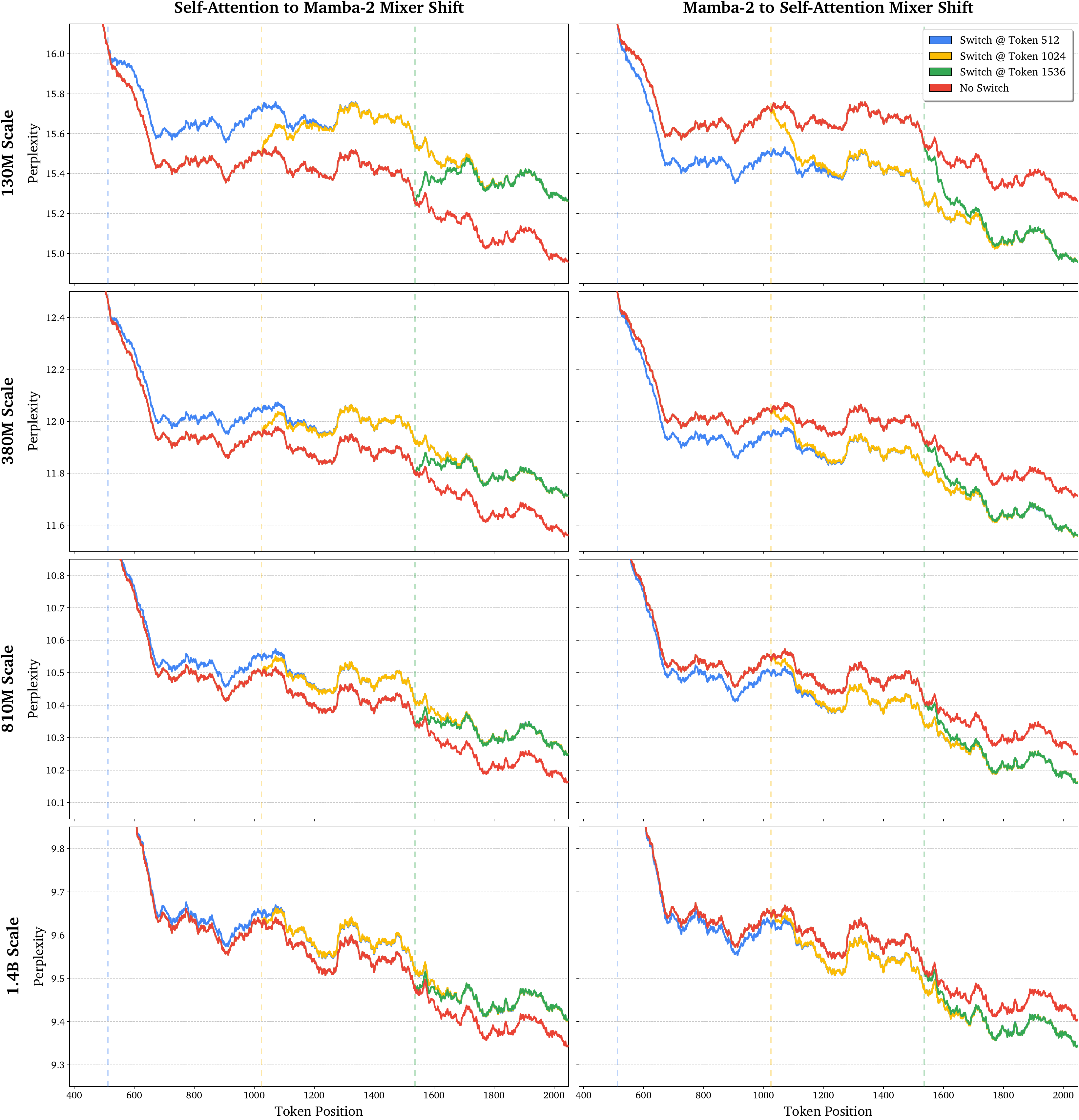}
    \caption{Smoothed perplexity across token index position for all chunked mixed-mode \antelope{}-TM models across scale.}
    \label{fig:mode-switch-all}
\end{figure}

\begin{figure}[h]
    \centering
    \includegraphics[width=\textwidth]{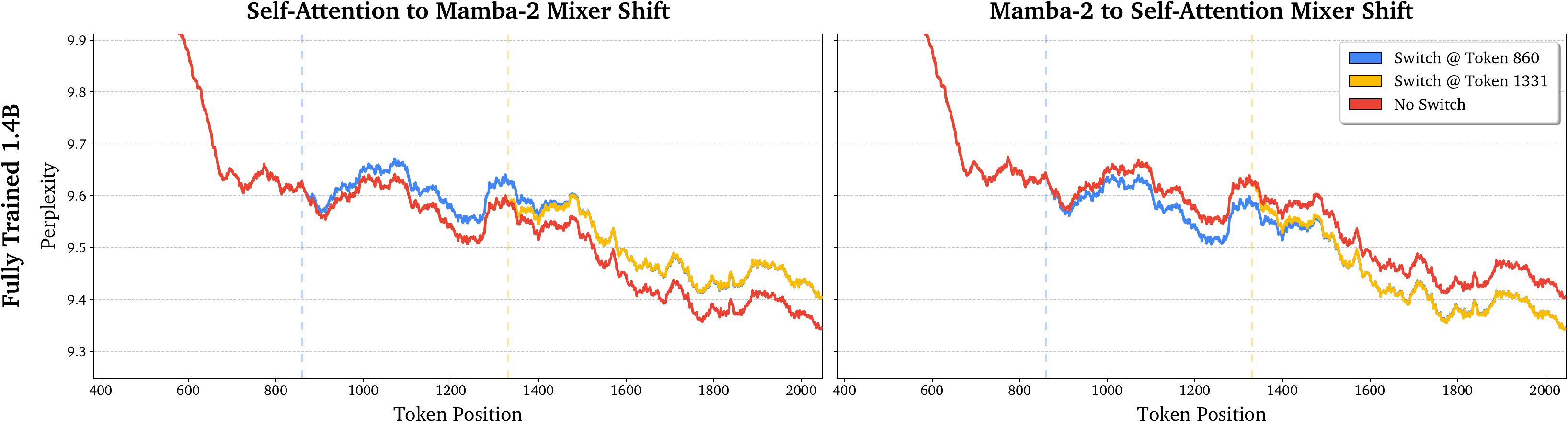}
    \caption{Smoothed perplexity across token index position at non-chunk boundaries for chunked mixed-mode pretrained \antelope{}-TM 1.4B model.}
    \label{fig:mode-switch-vary}
\end{figure}

\begin{figure}[h]
    \centering
    \includegraphics[width=\textwidth]{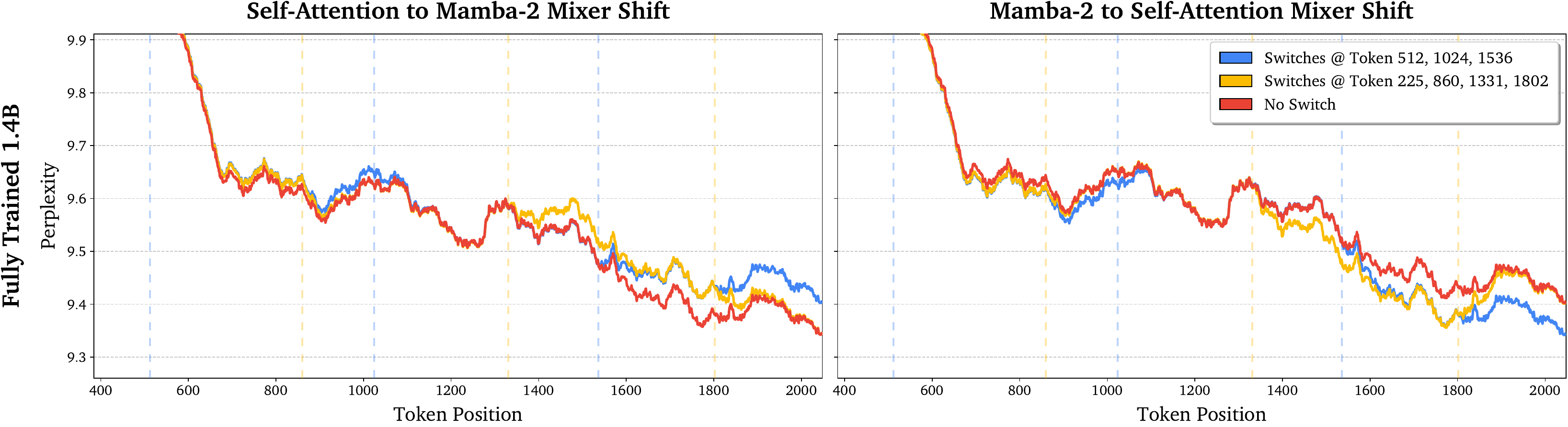}
    \caption{Smoothed perplexity across token index position for multiple switches for chunked mixed-mode pretrained \antelope{}-TM 1.4B model.}
    \label{fig:mode-switch-multi}
\end{figure}

We find that chunked mixed-mode training enables switching at all scales (\Cref{fig:mode-switch-all}), switching at non-chunk boundaries, i.e., non-multiples of 128 (\Cref{fig:mode-switch-vary}), and multiple switches within a single sequence (\Cref{fig:mode-switch-multi}).

\begin{figure}[h]
    \centering
    \includegraphics[width=\textwidth]{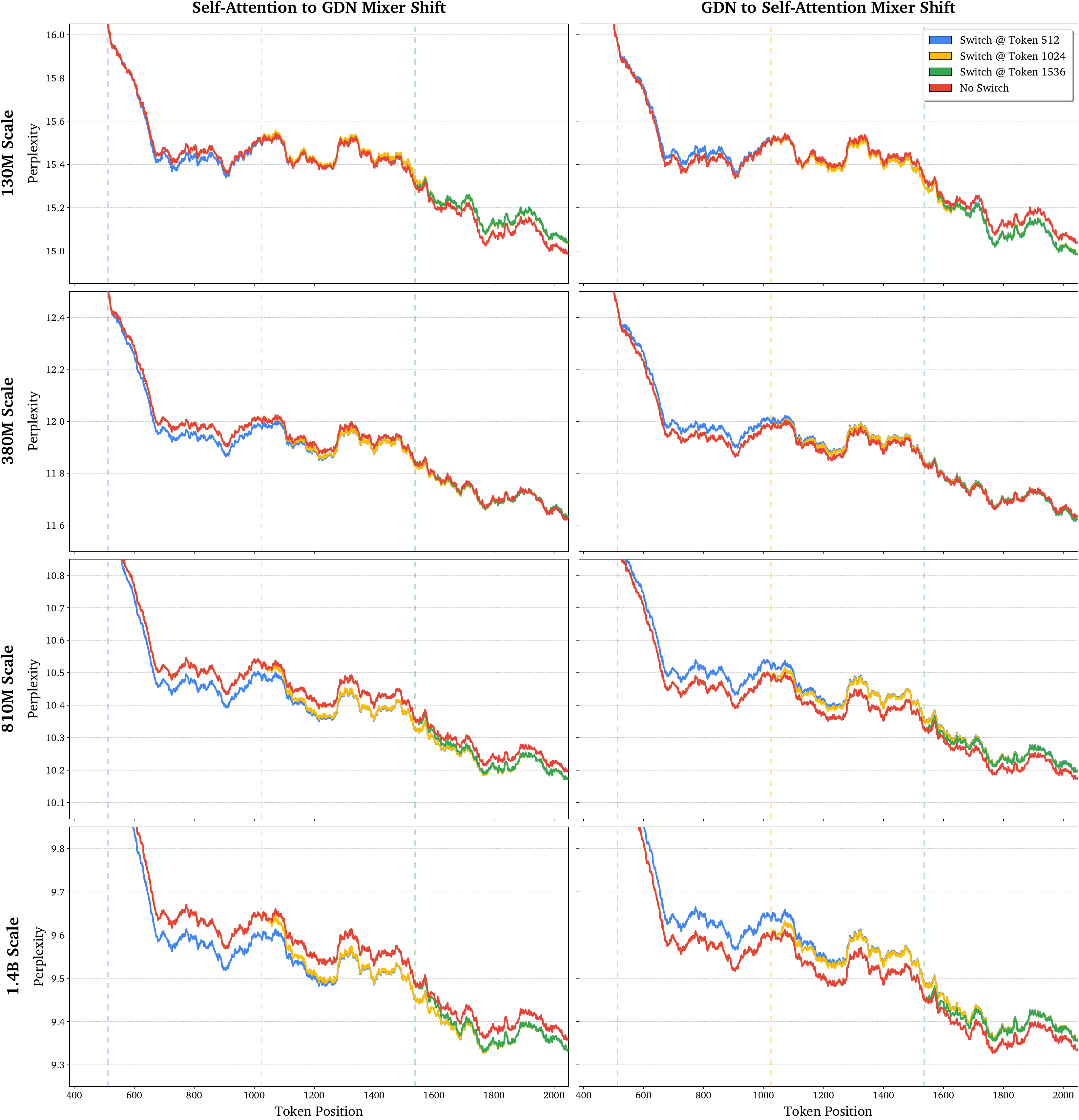}
    \caption{Smoothed perplexity across token index position for all chunked mixed-mode pretrained \antelope{}-TG models across scale.}
    \label{fig:mode-switch-all-gdn}
\end{figure}
\begin{figure}[h]
    \centering
    \includegraphics[width=\textwidth]{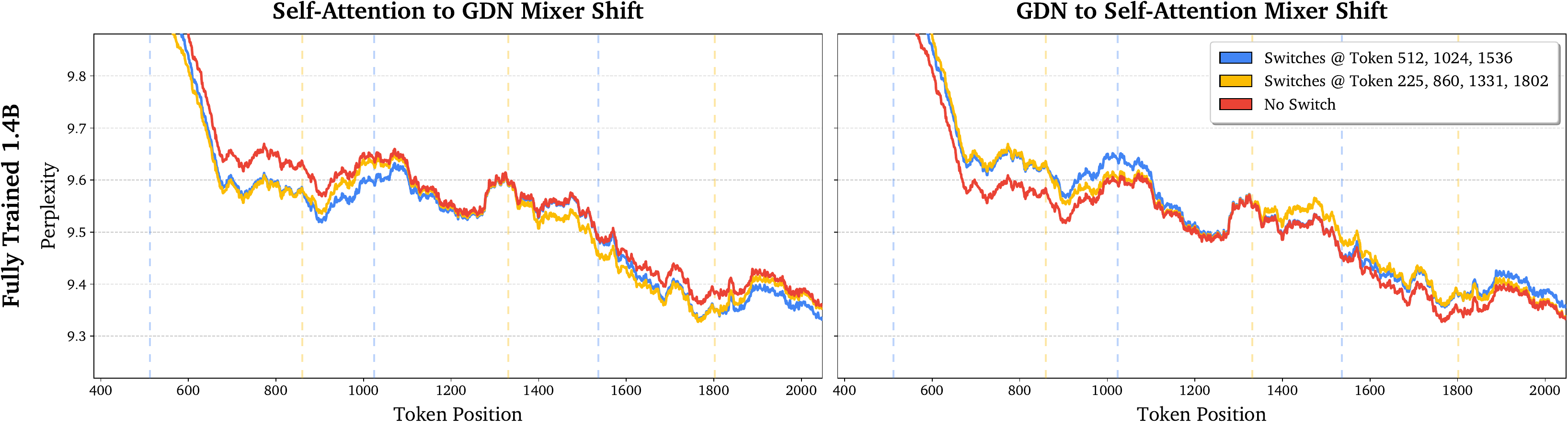}
    \caption{Smoothed perplexity across token index position for multiple switches for chunked mixed-mode pretrained \antelope{}-TG 1.4B model.}
    \label{fig:mode-switch-multi-gdn}
\end{figure}

\end{document}